\documentclass[10pt,a4paper]{article}
\usepackage{fontspec}
\newfontfamily\afuchinesefont{FandolSong-Regular.otf}[BoldFont=FandolSong-Bold.otf]
\newcommand{\afuagent}{Ant Afu}
\usepackage[english]{babel}
\usepackage[a4paper,margin=23mm]{geometry}
\usepackage{microtype}
\usepackage{amsmath,amssymb,mathtools}
\usepackage{booktabs,tabularx,array,longtable}
\usepackage{graphicx}
\usepackage[table]{xcolor}
\usepackage{enumitem}
\usepackage[numbers,sort&compress]{natbib}
\usepackage{caption}
\usepackage{tikz}
\usepackage{pgfplots}
\pgfplotsset{compat=1.18}
\usetikzlibrary{arrows.meta,positioning,calc,patterns,patterns.meta}
\usepackage{fancyhdr}
\usepackage{placeins}
\usepackage{float}
\usepackage{needspace}
\usepackage{titlesec}
\titleformat{\section}{\Large\bfseries\raggedright\hyphenpenalty=10000}{\thesection}{1em}{}
\titleformat{\subsection}{\large\bfseries\raggedright\hyphenpenalty=10000}{\thesubsection}{1em}{}
\titleformat{\subsubsection}{\normalsize\bfseries\raggedright\hyphenpenalty=10000}{\thesubsubsection}{1em}{}
\usepackage[colorlinks=true,allcolors=blue!45!black]{hyperref}

\definecolor{ink}{HTML}{17364D}
\definecolor{accent}{HTML}{287D8E}
\definecolor{muted}{HTML}{5C6773}
\definecolor{notebg}{HTML}{F0F4F7}
\definecolor{notefg}{HTML}{445B70}
\definecolor{reviewbg}{HTML}{EEF5F2}
\setlist{itemsep=2pt,topsep=4pt,leftmargin=*}
\renewcommand{\arraystretch}{1.06}
\definecolor{hcltablehead}{HTML}{DCD6EF}
\definecolor{hcltableband}{HTML}{ECE8F6}
\definecolor{hcltablerow}{HTML}{F4F2FA}
\definecolor{hcltableid}{HTML}{E6E1F2}
\definecolor{hcltexthead}{HTML}{E6EBEF}
\definecolor{hcltextband}{HTML}{F1F4F6}
\newcommand{\hcltablefont}{\fontsize{9}{10.5}\selectfont\renewcommand{\arraystretch}{1.00}\setlength{\tabcolsep}{3pt}}
\newcommand{\hcltexttablefont}{\hcltablefont}
\newcommand{\hcltextrowsep}{\addlinespace[1.5pt]}
\newcommand{\hcldensetablefont}{\fontsize{8.5}{9.6}\selectfont\renewcommand{\arraystretch}{1.00}\setlength{\tabcolsep}{2.5pt}\setlength{\arrayrulewidth}{0.2pt}}

\newcolumntype{L}[1]{>{\raggedright\arraybackslash}p{#1}}
\newcolumntype{Y}{>{\raggedright\arraybackslash}X}
\newcommand{\bench}{HealthCLBench}
\newcommand{\ind}{\mathbb{I}}
\hypersetup{pdftitle={HealthCLBench: Evaluating Context Learning in Health Conversations through Controlled Case Clusters},
  pdfauthor={Ting Yang, Zijing Hong, Chengjie Wang, Peng Wei, Chunxiao Guo, Yun Yue, Xurui Li, Jinqun Guan, Zixiao Chen},pdfsubject={Health context learning, related case clusters, and context ablation evaluation},
  pdflang={en-US}}

\title{\textbf{HealthCLBench:\\Evaluating Context Learning\\in Health Conversations\\through Controlled Case Clusters}}
\author{Ting Yang\quad Zijing Hong\quad Chengjie Wang\quad Peng Wei\quad Chunxiao Guo\\[2pt]
Yun Yue\quad Xurui Li\quad Jinqun Guan\quad Zixiao Chen\\[3pt]{\normalsize Ant Group}}
\date{}

\begin{document}
\maketitle
\begin{abstract}
Complex health conversations require answers grounded in individual information and currently valid evidence. We evaluate health context learning: acquiring, integrating, and applying relevant context during inference with fixed model parameters to answer health questions and respond appropriately to changes in facts, evidence, or task goals. To our knowledge, Chinese-language medical benchmarks have yet to systematically combine explicit contextual difficulties, controlled case variations, and member-specific medical criteria for open-ended answers. We introduce \textbf{\bench{}}, comprising \textbf{328 controlled clusters, 1,103 cases, and 4,931 rubric criteria} across medical consultation and everyday health management. Members share backgrounds while evidence, patient states, or task goals change; each has its own medical scoring requirements. A 70-person medical team annotates and reviews the simulated cases and their rubrics. Human review of 700 full-context responses from the original evaluation yields 90.5\% pooled agreement with the original automated judgments at the rubric-item level. We report full-context and query-only results for seven general-purpose models and \afuagent{}. Across the eight systems, scores with full context are 11.83 to 31.76 case points and 10.87 to 38.97 geometric cluster points higher than those with the current question alone. GPT-6 Astra leads both full-context metrics at 75.03 and 60.42, followed by Claude Opus 5.5 and \afuagent{}. Cross-condition differences describe score changes under the reported evaluation settings. Response comparisons reveal failures in checking historical information, expressing uncertainty when warranted, and ensuring recommendations respect required conditions. The resource supports model comparison and targeted follow-up evaluation.
\end{abstract}

% Generated by analysis/revision_20261110/render_assets.py.
\begin{figure}[H]\centering
\usetikzlibrary{patterns,patterns.meta}
\begingroup
\definecolor{hclHatchPrimary}{HTML}{6464A0}
\definecolor{hclHatchPrimaryEdge}{HTML}{515183}
\definecolor{hclHatchSecondary}{HTML}{F6C5BC}
\definecolor{hclHatchSecondaryEdge}{HTML}{D99D91}
\definecolor{hclHatchBackground}{HTML}{F7F7F9}
\tikzset{
  hclHatchPrimary/.style={fill=hclHatchPrimary,draw=hclHatchPrimaryEdge,line width=0.35pt,
    postaction={pattern={Lines[angle=45,distance=4.5pt,line width=0.23pt]},pattern color=white!75!hclHatchPrimary}},
  hclHatchSecondary/.style={fill=hclHatchSecondary,draw=hclHatchSecondaryEdge,line width=0.35pt,
    postaction={pattern={Hatch[angle=45,distance=4.5pt,line width=0.23pt]},pattern color=hclHatchSecondaryEdge!85!white}}
}

\begin{tikzpicture}\begin{axis}[
width=0.98\linewidth,height=46mm,ymin=0,ymax=100,ytick={0,20,40,60,80,100},
axis lines*=left,axis line style={draw=black!30,line width=0.35pt},tick style={draw=black!25},
tick label style={font=\fontsize{8}{9}\selectfont,text=black!85},label style={font=\fontsize{8.2}{9.2}\selectfont,text=black!85},ylabel={Score (0--100)},
axis background/.style={fill=hclHatchBackground},ymajorgrids=true,grid style={draw=white,line width=0.5pt},
legend style={font=\fontsize{8.4}{9.3}\selectfont,draw=none,fill=white,column sep=12pt,at={(0.5,1.02)},anchor=south},
xmin=-0.6,xmax=7.6,xtick={0,1,2,3,4,5,6,7},
xticklabels={{\shortstack{GPT-6\\Astra}},{\shortstack{Claude\\Opus 5.5}},{\afuagent{}},{\shortstack{Qwen3.8-\\27B}},{GLM-5.3},{Kimi K3},{\shortstack{DeepSeek-\\V4.1-Flash}},{\shortstack{Gemini\\3.8 Flash}}},
legend columns=2,bar width=10pt,clip=false]
\addplot[ybar,bar shift=0pt,forget plot,hclHatchPrimary] coordinates {(-0.18,75.030117860381)};
\node[anchor=south,font=\fontsize{7.5}{8.5}\selectfont,text=black!85] at (axis cs:-0.18,76.030117860381) {75.03};
\addplot[ybar,bar shift=0pt,forget plot,hclHatchSecondary] coordinates {(0.18,60.424731275358)};
\node[anchor=south,font=\fontsize{7.5}{8.5}\selectfont,text=black!85] at (axis cs:0.18,61.424731275358) {60.42};
\addplot[ybar,bar shift=0pt,forget plot,hclHatchPrimary] coordinates {(0.82,68.212339075249)};
\node[anchor=south,font=\fontsize{7.5}{8.5}\selectfont,text=black!85] at (axis cs:0.82,69.212339075249) {68.21};
\addplot[ybar,bar shift=0pt,forget plot,hclHatchSecondary] coordinates {(1.18,49.500240659763)};
\node[anchor=south,font=\fontsize{7.5}{8.5}\selectfont,text=black!85] at (axis cs:1.18,50.500240659763) {49.50};
\addplot[ybar,bar shift=0pt,forget plot,hclHatchPrimary] coordinates {(1.82,61.990072529465)};
\node[anchor=south,font=\fontsize{7.5}{8.5}\selectfont,text=black!85] at (axis cs:1.82,62.990072529465) {61.99};
\addplot[ybar,bar shift=0pt,forget plot,hclHatchSecondary] coordinates {(2.18,43.304163484191)};
\node[anchor=south,font=\fontsize{7.5}{8.5}\selectfont,text=black!85] at (axis cs:2.18,44.304163484191) {43.30};
\addplot[ybar,bar shift=0pt,forget plot,hclHatchPrimary] coordinates {(2.82,57.305403445150)};
\node[anchor=south,font=\fontsize{7.5}{8.5}\selectfont,text=black!85] at (axis cs:2.82,58.305403445150) {57.31};
\addplot[ybar,bar shift=0pt,forget plot,hclHatchSecondary] coordinates {(3.18,36.443405306297)};
\node[anchor=south,font=\fontsize{7.5}{8.5}\selectfont,text=black!85] at (axis cs:3.18,37.443405306297) {36.44};
\addplot[ybar,bar shift=0pt,forget plot,hclHatchPrimary] coordinates {(3.82,52.288032638259)};
\node[anchor=south,font=\fontsize{7.5}{8.5}\selectfont,text=black!85] at (axis cs:3.82,53.288032638259) {52.29};
\addplot[ybar,bar shift=0pt,forget plot,hclHatchSecondary] coordinates {(4.18,29.196484976580)};
\node[anchor=south,font=\fontsize{7.5}{8.5}\selectfont,text=black!85] at (axis cs:4.18,30.196484976580) {29.20};
\addplot[ybar,bar shift=0pt,forget plot,hclHatchPrimary] coordinates {(4.82,48.457162284678)};
\node[anchor=south,font=\fontsize{7.5}{8.5}\selectfont,text=black!85] at (axis cs:4.82,49.457162284678) {48.46};
\addplot[ybar,bar shift=0pt,forget plot,hclHatchSecondary] coordinates {(5.18,24.771133927628)};
\node[anchor=south,font=\fontsize{7.5}{8.5}\selectfont,text=black!85] at (axis cs:5.18,25.771133927628) {24.77};
\addplot[ybar,bar shift=0pt,forget plot,hclHatchPrimary] coordinates {(5.82,45.714315503173)};
\node[anchor=south,font=\fontsize{7.5}{8.5}\selectfont,text=black!85] at (axis cs:5.82,46.714315503173) {45.71};
\addplot[ybar,bar shift=0pt,forget plot,hclHatchSecondary] coordinates {(6.18,23.669110632779)};
\node[anchor=south,font=\fontsize{7.5}{8.5}\selectfont,text=black!85] at (axis cs:6.18,24.669110632779) {23.67};
\addplot[ybar,bar shift=0pt,forget plot,hclHatchPrimary] coordinates {(6.82,42.774052583862)};
\node[anchor=south,font=\fontsize{7.5}{8.5}\selectfont,text=black!85] at (axis cs:6.82,43.774052583862) {42.77};
\addplot[ybar,bar shift=0pt,forget plot,hclHatchSecondary] coordinates {(7.18,21.792379455611)};
\node[anchor=south,font=\fontsize{7.5}{8.5}\selectfont,text=black!85] at (axis cs:7.18,22.792379455611) {21.79};
\addlegendimage{area legend,hclHatchPrimary}
\addlegendentry{Case score (diagonal)}
\addlegendimage{area legend,hclHatchSecondary}
\addlegendentry{Geometric cluster score (crosshatch)}
\end{axis}\end{tikzpicture}\endgroup
\caption{Full-context scores for eight systems on 1,103 cases in 328 clusters. Both metrics give the same ranking. Detailed scores are in Table~\ref{tab:overall-results}.}
\label{fig:model-scores-grouped}
\end{figure}
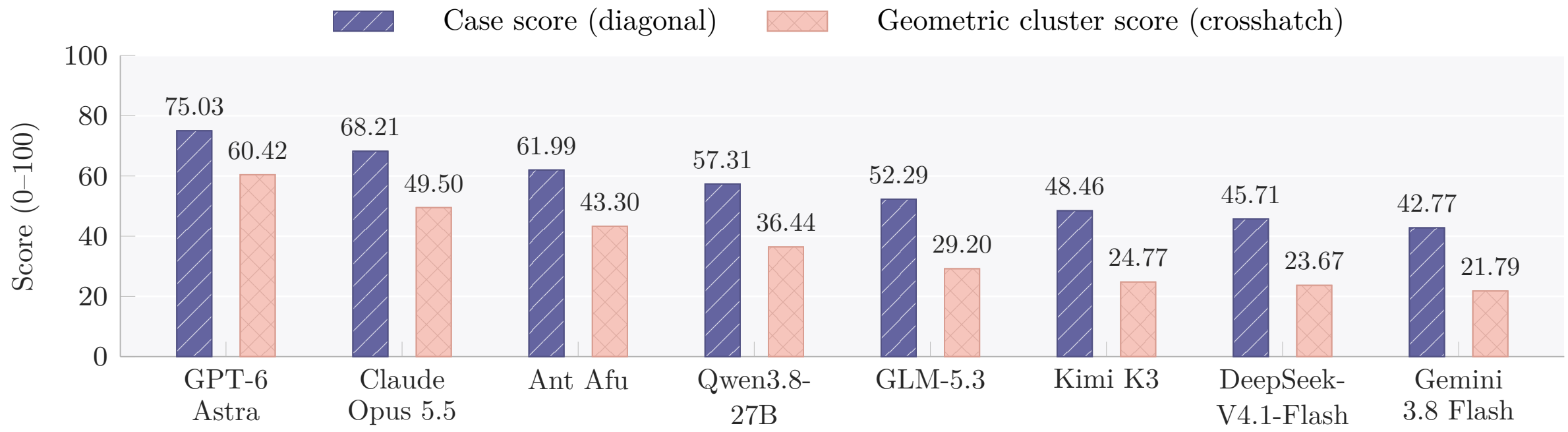
\section{Introduction}
\label{sec:introduction}

A current question in a health conversation often becomes meaningful only in light of earlier dialogue and the individual's health background. Users may describe symptoms colloquially, supply test results across turns, or correct earlier information. A reply such as ``the first time'' requires recovering the symptom and event under discussion; newly reported cough and sore throat must be interpreted alongside earlier nasal congestion and headache. A model must identify whose information it is, when it applies, and whether it has been updated before completing an explanation, assessment, decision, triage, or health-management task. Evaluation should therefore examine whether an answer meets the current need and whether its conclusions are supported by the available evidence. It should also assess how explanations, judgments, and recommendations adapt when evidence, patient state, or task goals change.

Existing benchmarks establish important foundations. HealthBench combines realistic health-interaction features with physician-authored rubrics to evaluate open-ended responses, including context awareness\citep{arora2025healthbench}. Among Chinese-language resources, CMB-Clin uses case descriptions and preceding questions and answers for multi-turn diagnostic and treatment tasks\citep{wang2024cmb}; ClinConsensus evaluates open-ended responses through case-specific rubrics and physician-calibrated coverage thresholds\citep{zheng2026clinconsensus}; and MedBench v4 spans text, multimodal, and agent tasks, including long conversations and extended clinical documents\citep{ding2025medbenchv4}. These resources support evaluation of medical knowledge application, specific response requirements, and clinical workflows. MedEinst and MedPIC-Bench further use controlled case variations and paired metrics to examine how differential diagnosis and medication rules respond to patient information\citep{chen2026medeinst,hou2026medpic}.

To our knowledge, no existing Chinese-language medical benchmark systematically integrates explicit characterization of contextual difficulties, controlled case variations within shared health settings, open-ended responses, and member-specific medical evaluation criteria in a unified resource. Table~\ref{tab:resource-comparison} summarizes this gap and our positioning.

To address this gap, we introduce \textbf{\bench{}}\footnote{Dataset: \url{https://huggingface.co/datasets/AQ-MedAI/HealthCLBench}.}, a Chinese-language benchmark for health context learning comprising \textbf{328 clusters, 1,103 cases, and 4,931 rubric criteria}. Each case combines a health setting and a current question, spanning medical consultation, diet, exercise, and everyday health management. Initial tasks are constructed from simulated health dialogues. Controlled expansion, termed Diffusion, retains shared backgrounds, records evidence, state, or task-goal changes, and assigns independent inputs and scoring requirements to each member. Four context-difficulty categories with 14 subtypes identify requirements for relation tracking, expression understanding, evidence integration, and assessment of conflicting information. A team of 70 participants, all holding at least a medical bachelor's degree and 90\% licensed physicians, conducts full annotation followed by sampled quality control and acceptance checks. They check medical plausibility, subject and time attribution, and alignment among inputs, tasks, and criteria.

The benchmark focuses on \emph{health context learning}: acquiring, integrating, and applying relevant information supplied in the context during inference with fixed model parameters to answer health questions and respond appropriately to changes in facts, evidence, or task goals. This definition emphasizes combining individual facts in the context with applicable medical knowledge, evaluated through task performance on independently answered cases.

The protocol provides two complementary views. Full-context versus query-only comparisons (Full--Query) retain the question and rubrics of the same case, comparing task scores with and without supplementary background. Within-cluster comparisons use each member's inputs and requirements to assess performance across the related tasks within a cluster. Each member is answered independently. Low-scoring members reduce the geometric cluster score even when other members score well; specific failures can then be traced to inputs, responses, and rubric items. The parotid example in Section~\ref{sec:design} and its results in Section~\ref{sec:paired-main} demonstrate this use; Appendix~\ref{sec:paired-responses} retains the complete eight-system scores.

Our contributions are:
\begin{enumerate}
  \item \textbf{A benchmark of controlled case clusters for complex health contexts.} Medically reviewed clusters combine complex input conditions, controlled case relations, and member-specific requirements for open-ended responses, supporting evaluation of contextual answering across related health situations.
  \item \textbf{Member-specific medical scoring criteria.} Rubrics calibrated through full medical annotation and sampled quality control and acceptance checks support item-level checks of critical facts, necessary explanations, recommendations, and whether conclusions stay within what the evidence supports. They align task variations with response requirements.
  \item \textbf{Full-context and query-only reference results for eight systems.} Full--Query comparisons, case quality, joint cluster performance, subgroup results, and rubric-level grading evidence support subsequent model comparison, localization of weak tasks, and selection of targeted regression evaluations.
\end{enumerate}

Under full context, the eight systems have the same ordering on both metrics, led by GPT-6 Astra, Claude Opus 5.5, and \afuagent{}. Across the eight systems, Full--Query differences range from 11.83 to 31.76 case points and from 10.87 to 38.97 cluster points. These differences describe performance changes under the reported conditions (Section~\ref{sec:context-ablation}). Section~\ref{sec:experiments} further examines failures in checking historical information, expressing uncertainty when warranted, and ensuring recommendations respect required conditions, informing subsequent capability evaluation and regression-task selection.

\begin{table}[htbp]
\centering\hcltexttablefont
\caption{Principal task organization and scoring of related resources, as described in the cited versions.}
\label{tab:resource-comparison}
\begin{tabular}{@{}L{26mm}L{62mm}L{70mm}@{}}
\toprule
\rowcolor{hcltexthead}
\textbf{Resource} & \textbf{Tasks and relations} & \textbf{Responses and scoring} \\
\midrule
HealthBench\newline\citep{arora2025healthbench}
& Health conversations; includes context awareness
& Open-ended final responses; conversation-specific rubrics \\
\hcltextrowsep
CMB-Clin\newline\citep{wang2024cmb}
& Chinese multi-turn clinical tasks;\newline{}74 cases, 208 questions
& Open answers; fluency, relevance, completeness, and medical proficiency \\
\hcltextrowsep
ClinConsensus v6\newline\citep{zheng2026clinconsensus}
& 2,500 Chinese open-ended clinical cases
& 30 binary criteria per case; physician-calibrated coverage thresholds \\
\hcltextrowsep
MedBench v4\newline\citep{ding2025medbenchv4}
& Chinese medical tasks; includes long conversations and documents
& Text, multimodal, and agent tracks; task-specific scoring \\
\hcltextrowsep
MedEinst\newline\citep{chen2026medeinst}
& Diagnostic pairs with changed discriminating evidence
& Diagnostic labels; case and paired accuracy, bias trap rate \\
\hcltextrowsep
MedPIC-Bench\newline\citep{hou2026medpic}
& Medication-safety pairs with changed patient information
& Multiple choice; option exact match and pair accuracy \\
\midrule
\rowcolor{hcltextband}
\textbf{This work}
& 1,103 Chinese tasks, 328 controlled clusters;\newline{}input difficulties and evidence/state/goal changes
& Open answers, member rubrics; case and geometric cluster scores; Full--Query comparisons \\
\bottomrule
\end{tabular}
\end{table}

\FloatBarrier

\section{Benchmark Design and Data Distribution}
\label{sec:design}

\subsection{Related Clusters and Task Definition}
The benchmark assesses health context learning through context-dependent answers to complex medical and health questions, connecting cases through shared health contexts and controlled variations. Each case may concern medical consultation, diet, exercise, or everyday health management; it need not involve illness or a confirmed diagnosis. An individual case is written as
\begin{equation}
 x_i=(C_i,q_i,R_i,z_i),\qquad a_i=M(C_i,q_i),
\end{equation}
where $q_i$ is the current question, including its explicitly stated facts, $C_i$ contains the dialogue history and relevant health materials supplied in addition to $q_i$, $R_i$ is the set of task-specific rubrics, and $z_i$ contains task and content labels. $M$ denotes the evaluated system, whose response-generating model weights remain fixed during inference; $a_i=M(C_i,q_i)$ is its full-context response, which the rubrics evaluate. For the same case, $q_i$ is held fixed across the full-context and query-only conditions; query-only inputs remove the remaining history and health materials. Both conditions use the same rubrics and scoring function, and neither provides scoring criteria or reference answers to the evaluated system. Section~\ref{sec:context-ablation} compares the corresponding cases.

An evaluation cluster is defined as
\begin{equation}
 \mathcal C_g=\{x_{g,1},\ldots,x_{g,n_g}\},\qquad n_g=|\mathcal C_g|.
\end{equation}
Here $g$ indexes clusters and $j\in\{1,\ldots,n_g\}$ indexes members within a cluster; $x_{g,j}$ is the same case as $x_i$ under the global index $i=i(g,j)$. Members share a context and are connected by explicitly describable changes in evidence, state, or task. All members have equal status: any member can serve as a reference in a local comparison, and neither cluster membership nor scoring depends on a particular member serving as the construction starting point. The same disease may appear in multiple clusters, and a cluster may contain different task labels. Clusters are organized by shared contexts and explicitly defined relationships between controlled variations.

\paragraph{How a cluster works: a three-member example.}
Cluster B in Appendix~\ref{sec:diffusion-examples} contains three members sharing a history of outer lower-front gum pain and swelling and left-sided palate and swallowing pain. Table~\ref{tab:cluster-walkthrough} links each member's current information and response goal to checkable rubric requirements. B1 and B2 vary in both test evidence and question scope, so their comparison focuses on shared management requirements. B3 asks about department choice using the shared symptoms alone; neither test result from the other members is available to it. Members are answered and scored separately before aggregation (Section~\ref{sec:scoring}); Table~\ref{tab:case-b-scores} reports complete scores for the seven general-purpose models and \afuagent{}, allowing readers to trace lost credit to individual members.
\begin{table}[H]
\centering\hcltexttablefont
\caption{All three members of Cluster B, linking current information, response goals, and selected rubric requirements. Table~\ref{tab:case-b} gives the questions and summarized requirements. All members retain the same symptom history.}
\label{tab:cluster-walkthrough}
\begin{tabularx}{\linewidth}{@{}L{43mm}L{25mm}Y@{}}
\toprule
\rowcolor{hcltexthead}
\textbf{Member and current information} & \textbf{Response goal} & \textbf{Checkable rubric requirements (selected)} \\
\midrule
\textbf{B1}: Adds parotid aspirate containing pus and a positive \textit{Staphylococcus aureus} culture. & How to manage the condition. & P0: Identify suppurative bacterial parotid infection and distinguish it from viral mumps; prioritize clinician-directed antibiotics informed by culture or susceptibility. P1: Recommend follow-up for recurrent pus and possible further drainage. \\
\hcltextrowsep
\textbf{B2}: Adds positive mumps-virus IgM. & Explain cause, management, and possible consequences. & P1: Treat IgM as supporting evidence, not standalone confirmation, and recommend clinical assessment; cover transmission precautions, supportive care, and complications; retain possible independent gum disease. \\
\hcltextrowsep
\textbf{B3}: Adds neither test result; asks about care if symptoms persist until Friday. & Which department to visit first. & P0: Choose dentistry first if both symptoms persist. P1: Choose otolaryngology (ENT) if mainly throat or palate pain remains; seek earlier care for fever, marked facial swelling, or suddenly worsening pain before Friday. \\
\bottomrule
\end{tabularx}
\par\vspace{3pt}
{\fontsize{8.5}{9.6}\selectfont\raggedright P0 denotes critical requirements; P1 denotes other requirements essential to task completion. See Section~\ref{sec:scoring} for the full scoring rules.\par}
\end{table}

Under this operational definition, health context learning is assessed through observable response requirements: attributing information to the correct individual and time, interpreting input expressions, integrating dispersed evidence, assessing conflicting information, and applying currently valid information. Members are answered independently to examine appropriate responses under different facts, evidence, or task conditions; this does not constitute training across cases. Rubrics are defined from the complete task materials and applicable medical and health knowledge, including requirements to express uncertainty appropriately when information is insufficient. Case scores measure satisfaction of these requirements in individual responses, and cluster evaluation examines joint performance across related questions. Grouping by medical scenario, context difficulty, and other labels characterizes differences across tasks, while full-context versus query-only comparisons reveal task-performance differences with and without supplementary background.

\subsection{Labeling Framework and Context Difficulty Definitions}
\label{sec:taxonomy}
The benchmark uses four labeling dimensions: medical scenario, context difficulty, disease, and department. Medical scenarios describe the task to be completed, context difficulties describe challenges in the input structure, and disease and department labels describe content coverage. Labels are defined from task inputs rather than inferred from evaluated model responses or errors. The table pairs definitions with condensed summaries of existing cases, showing how users can select scenarios by task and difficulties by input requirements. These summaries guide reading; evaluation uses the complete inputs and their rubrics.

\paragraph{Medical scenarios.}
Each case is assigned a scenario according to the main intent of the current question and the necessary background. This classification covers both clinical consultation and everyday health management. Reports, images, medications, and symptoms are evidence or content types, not separate task scenarios.
\paragraph{Context difficulties.}
Context difficulties comprise four categories and 14 subcategories. A case can have several difficulties, with the most decisive for completing the current task designated as its primary label. Cases outside these four defined categories are grouped as Other Difficulties. Table~\ref{tab:difficulties} defines the four categories, and Table~\ref{tab:difficulty-definitions} in Appendix~\ref{sec:difficulty-details} specifies the subcategories and their boundaries.
\begin{table}[H]
\centering\hcltexttablefont
\caption{Five medical scenarios and four context difficulty categories, with definitions and summaries of actual inputs. S1--S5 and D1--D4 map to the primary scenario and difficulty in each row, respectively. Summaries are paraphrases and do not replace complete inputs.}
\label{tab:scenarios}\label{tab:difficulties}
\begin{tabularx}{\linewidth}{@{}L{29mm}L{53mm}Y@{}}
\toprule
\rowcolor{hcltexthead}
\textbf{Category} & \textbf{Definition / primary task} & \textbf{Example summary: input and task} \\
\midrule
\rowcolor{hcltextband}
\multicolumn{3}{@{}l@{}}{\textbf{Medical scenarios (S1--S5)}}\\*
Medical Explanation & Explain diseases, medications, or medical concepts in relation to the request. & \textbf{S1}: Given C-reactive protein of 71.32 mg/L, explain the value and whether it alone identifies a specific disease. \\
\hcltextrowsep
Health Assessment & Integrate symptoms, history, and reports to assess individual health status. & \textbf{S2}: Earlier turns report about three days of nasal congestion, concurrent headache, both clear and yellow discharge, and no allergic rhinitis history. The current turn adds cough and sore throat for continued assessment. \\
\hcltextrowsep
Clinical Decision-Making & Address medication, treatment, discontinuation, switching, or choices among plans. & \textbf{S3}: A breastfeeding user finds codeine phosphate on a cough syrup label and asks whether it can be taken. \\
\hcltextrowsep
Care Triage & Assess risk and determine whether, where, and in which department to seek care. & \textbf{S4}: Outer lower-front gum pain and swelling and left-sided swallowing pain persist. The user asks which department to visit first if still unwell on Friday. \\
\hcltextrowsep
Health Management & Support diet, exercise, sleep, chronic care, and follow-up. & \textbf{S5}: After five sets of 20 squats on Sunday, leg pain and tenderness on days two and three restrict sitting, squatting, and walking. The user asks whether to continue lower-body training. \\
\midrule
\rowcolor{hcltextband}
\multicolumn{3}{@{}l@{}}{\textbf{Context difficulties (D1--D4)}}\\*
Complex Context & Interrelated people, topics, references, ellipsis, task requirements, or changing facts require relationship tracking and identification of current information. & \textbf{D1}: After describing tenderness below the right ribs following an oily dinner, the user answers only that this is the first time. The omitted symptom and event must be recovered. \\
\hcltextrowsep
Noisy Context & Facts remain recoverable despite colloquial language, input noise, or irrelevant material. Normalization or removal of distractions preserves the facts and answer. & \textbf{D2}: The user writes ``{\fontspec{FandolSong-Regular.otf}类似于豆豆的包}'' (literally, ``bumps like little beans'') for many longstanding back bumps. Interpret the colloquial wording without treating it as a confirmed diagnosis. \\
\hcltextrowsep
Fragmented Context & Complementary evidence spans turns or time points. Records, event relationships, or cross-turn facts must be integrated; one fragment is insufficient. & \textbf{D3} (also S2): The current turn adds only cough and sore throat. Earlier turns separately describe nasal congestion lasting about three days, headache, and the characteristics of nasal discharge. \\
\hcltextrowsep
Conflicting Context & Facts, applicability conditions, or action recommendations contain unresolved inconsistencies not explained by correction, state change, or superseding information. & \textbf{D4}: CA19-9 rose from 39 last year to 48 last month. Earlier discussion distinguished elevation from cancer, whereas the user now cites a video equating it with pancreatic cancer; these interpretations must be distinguished. \\
\bottomrule
\end{tabularx}
\end{table}

An explicit correction or state change is a dynamic context update, not automatically a conflict merely because earlier and later statements differ. Input noise requires the intended meaning to be uniquely recoverable from the context; missing decisive information is not recoverable noise. Health and medical knowledge constitute the domain background and are not a separate context difficulty category.

\paragraph{Diseases and departments.}
Diseases and departments use multilabel vocabularies with Other Diseases and Other Departments categories. Disease labels describe current conditions, medical history, family members' conditions, and suspected or differential diagnoses actually involved in the input; explicitly denied or ruled-out conditions are excluded. Department labels describe relevant departments and those responsible for the corresponding diseases. Both label types characterize case content rather than confirmed diagnoses, prevalence, or actual clinical visits. A case can have several labels or none of the corresponding type.

\subsection{Dataset Scale and Coverage}
\label{sec:distribution}
\bench{} contains 328 clusters comprising 1,103 cases. Each cluster has one or three to five members, and all cases contribute to both case-level and cluster-level evaluation. Table~\ref{tab:dataset} summarizes the scale and structure of the benchmark.

Cluster size reflects both the expansion design and subsequent selection. Diffusion uses two directions, evidence changes and task changes, while unexpanded cases remain as singleton clusters. In the current release, retained multimember clusters contain 3--5 cases, with no two-member clusters. An individual cluster may retain variants from only one direction or omit its original reference case; cluster definitions and scores use all members retained in the final benchmark.
% Overall statistics and exact label distributions, placed together in Section 3.
\begin{table}[H]
\centering\hcltablefont
\caption{Benchmark statistics and label distributions. Quartiles use cases for rubric counts and dialogue turns, and clusters for cluster size. Scenarios and primary difficulties each form a mutually exclusive partition; percentages use all 1,103 cases as the denominator. Other Difficulties covers cases outside the four defined categories.}
\label{tab:dataset}\label{tab:label-distribution}
\begin{minipage}[t]{0.485\linewidth}\vspace{0pt}
\textbf{(a) Overall statistics}\par\vspace{2pt}
\rowcolors{2}{hcltablerow}{white}
\begin{tabularx}{\linewidth}{@{}Yr@{}}
\toprule
\rowcolor{hcltablehead}
\textbf{Statistic} & \textbf{Value} \\
\midrule
Clusters & 328 \\
Cases & 1,103 \\
Members per cluster: median [Q1,Q3] & 3.5 [3,4] \\
Members per cluster: minimum--maximum & 1--5 \\
Rubric criteria & 4,931 \\
P0 / P1 / P2 criteria & 1,028 / 3,648 / 255 \\
Criteria per case: median [Q1,Q3] & 4 [3,5] \\
Criteria per case: minimum--maximum & 2--14 \\
User turns: median [Q1,Q3] & 4 [2,8] \\
User turns: minimum--maximum & 1--50 \\
\bottomrule
\end{tabularx}
\end{minipage}\hfill%
\begin{minipage}[t]{0.485\linewidth}\vspace{0pt}
\textbf{(b) Medical scenarios and primary difficulties}\par\vspace{2pt}
\rowcolors{2}{hcltablerow}{white}
\begin{tabularx}{\linewidth}{@{}Yrr@{}}
\toprule
\rowcolor{hcltablehead}
\textbf{Label} & \textbf{Cases} & \textbf{Share (\%)} \\
\midrule
Scenario: Medical Explanation & 41 & 3.72 \\
Scenario: Health Assessment & 498 & 45.15 \\
Scenario: Clinical Decision-Making & 273 & 24.75 \\
Scenario: Care Triage & 127 & 11.51 \\
Scenario: Health Management & 164 & 14.87 \\
\midrule
Primary difficulty: Complex Context & 545 & 49.41 \\
Primary difficulty: Noisy Context & 65 & 5.89 \\
Primary difficulty: Fragmented Context & 305 & 27.65 \\
Primary difficulty: Conflicting Context & 81 & 7.34 \\
Primary difficulty: Other Difficulties & 107 & 9.70 \\
\bottomrule
\end{tabularx}
\end{minipage}
\end{table}

\paragraph{Dialogue turns.}
Dialogue turns are the user utterances in each case, including the current question. The unit of analysis is the case, so members sharing dialogue backgrounds are counted separately. The median is 4 turns, the interquartile range is 2--8, the mean is 7.22, and the range is 1--50. There are 213 cases (19.3\%) with more than 10 turns and 18 (1.6\%) with more than 40, forming a long right tail. Figure~\ref{fig:dialogue-turn-distribution}(a) shows the distribution. The 157 single-turn cases may also contain health background needed to complete the task.

\paragraph{Question types and primary difficulties.}
Health Assessment contains 498 cases (45.1\%) and Clinical Decision-Making 273 (24.8\%), together accounting for 69.9\%. Health Management, Medical Explanation, and Care Triage account for 14.9\%, 3.7\%, and 11.5\%, respectively. Among primary difficulties, Complex Context contains 545 cases, Fragmented Context 305, Conflicting Context 81, and Noisy Context 65. The remaining 107 cases (9.7\%) are grouped as Other Difficulties, covering cases outside these four defined categories. Figure~\ref{fig:case-distribution-pies}(b,c) shows these two distributions.

\paragraph{Co-occurring difficulties.}
Deduplicated within each case, multilabel coverage is 765 cases (69.4\%) for Complex Context, 679 (61.6\%) for Fragmented Context, 359 (32.5\%) for Noisy Context, and 181 (16.4\%) for Conflicting Context. A case may appear in several groups, so counts cannot be added to obtain the number of cases. The most frequent subcategories are evidence fragmented across turns (532 cases), reference and ellipsis dependence (510), and dynamic context updates (425), reflecting coverage of evidence integration, relationship resolution, and state updates.
\input{figures/dataset_distributions}

\paragraph{Disease and department coverage.}
Disease and department labels provide content-based retrieval dimensions and use multilabel statistics. Appendix~\ref{sec:benchmark-details} reports detailed counts, their denominators, and category-occurrence plots. These distributions do not represent confirmed diagnoses, prevalence, or actual clinical visits.

\paragraph{Multidimensional coverage and diversity.}
The benchmark spans five question types, four context difficulty categories and their 14 subcategories, with diseases, departments, and dialogue lengths providing additional dimensions of health needs. Tasks include consultations about tests, symptoms, and treatment, as well as diet, exercise, and everyday health management. Input structures require relationship tracking, interpretation of expressions, integration of dispersed evidence, and assessment of conflicting information. Together, these dimensions support analyses of model performance across tasks and contexts. Categories are not evenly distributed, so we report both overall and subgroup results to characterize the scope and composition of coverage.
\FloatBarrier

\section{Data Production and Human Medical Review}
\label{sec:production}

\begin{figure}[H]
\centering
\includegraphics[width=0.84\linewidth]{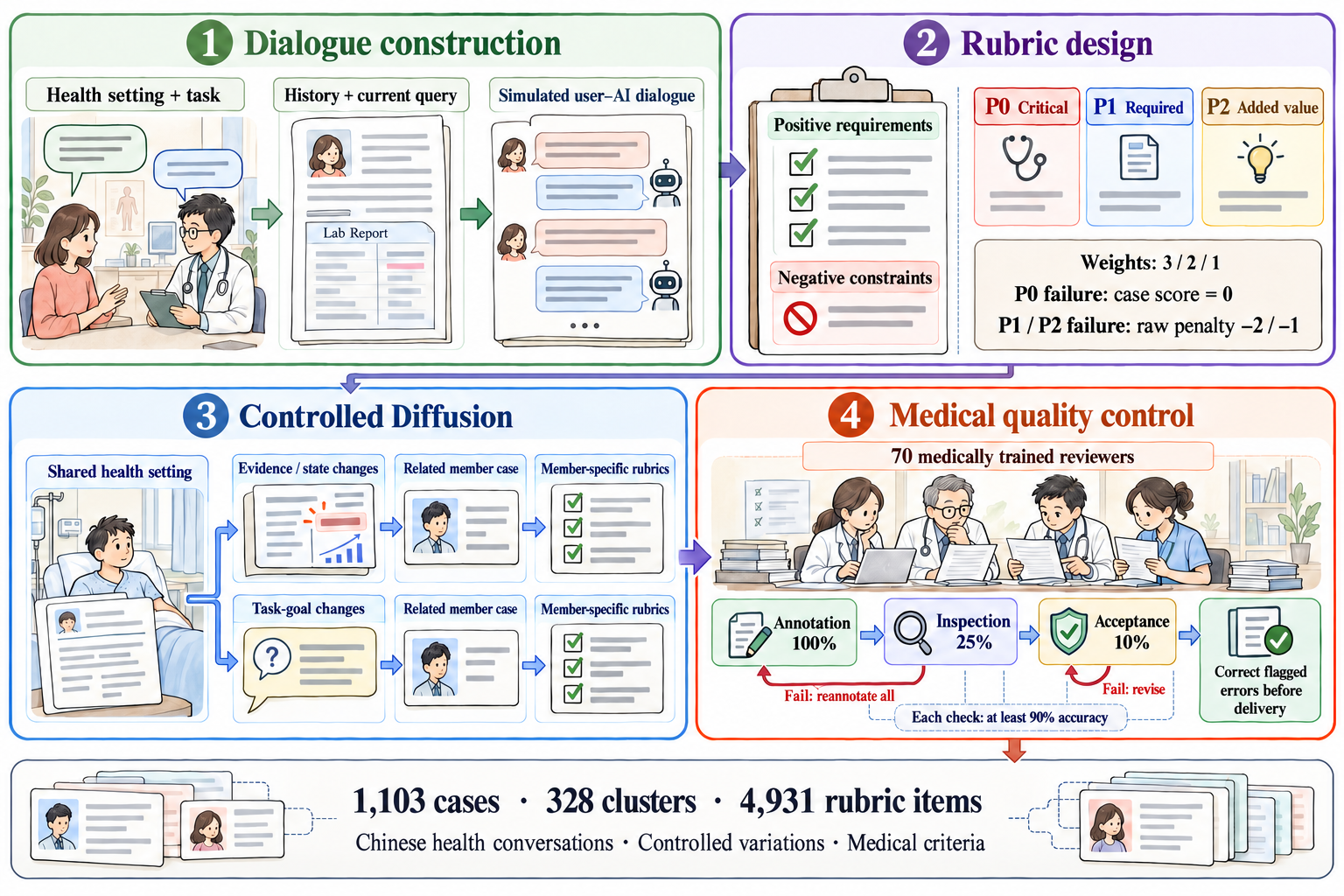}
\caption{Data production and medical quality control: dialogue construction, rubric design, controlled variation, and medical review. Full annotation is followed by sampled inspection and acceptance, with errors corrected before delivery. Reviewer qualifications, sampling, and rework rules are detailed in this section; response scoring is defined in Section~\ref{sec:scoring}. People and records are schematic.}
\label{fig:production}
\end{figure}

The initial cases in \bench{} are generated from predesigned simulated dialogues between users and AI. We first specify health settings and dialogue objectives, simulate user exchanges about health, and use a health dialogue system to generate complete conversations. We then produce rubrics, apply Diffusion, and conduct medical review. This process organizes medical question-answering tasks that require understanding, integrating, and applying contextual information. Scenario, difficulty, and clinical-content labels describe the benchmark's tasks and coverage. Figure~\ref{fig:production} shows the main steps.

% Hand-drawn artwork generated with image_gen; process and labels manually checked.

\subsection{Dialogue Design and Simulated Generation}
Health settings are designed around the labeling framework, specifying the simulated user's background, main concern, dialogue objective, and context difficulties to examine. Input structure follows the task requirements: a single-turn consultation can state the necessary health conditions directly, while multi-turn simulation introduces relevant information progressively, linking historical exchanges to the current question. A health dialogue system generates the AI side of the conversation, and initial evaluation cases are derived from the designed conversations. Each case includes the health facts and materials needed for the current task, allowing response requirements to examine information acquisition, relational integration, and task-specific application together.

\subsection{Rubric Production and Label Recording}
Rubrics are generated automatically from the complete context and current task to specify the key requirements a response must satisfy, with importance levels P0, P1, and P2. Rubric items should focus on the current question, be independently judgeable, and agree with the simulated setting's facts and applicable medical and health knowledge. These task-specific requirements test whether relevant information is correctly applied in the response. Scenario, difficulty, and clinical labels describe input properties; they do not serve as independent evidence for generating correct answers or scoring requirements and are separate from target model responses. Both full-context and query-only conditions use the scoring protocol in Section~\ref{sec:scoring}.

\subsection{Diffusion Expansion}
Starting from a task in a shared health setting, we identify clues relevant to the current health question and construct related variants by varying individual states, measurements, pathogen-related evidence, drug ingredients, evidence attribution, or the current task. Corresponding rubrics are generated for each task. Section~\ref{sec:diffusion} presents the controlled expansion method and the medical rationale for its clinical examples. Relationships within a cluster record shared background and evidence changes for human review and cluster-level evaluation.

\subsection{Medical Annotation, Sampled Quality Control, and Acceptance}
The medical review team comprises \textbf{70 participants}, all holding a bachelor's degree or higher in a medical discipline; \textbf{90\% are licensed physicians}. Quality control has three stages: annotation of the full dataset, a 25\% quality-control sample, and a 10\% acceptance sample. Both sampling stages calculate correctness at the rubric-item level: the number of inspected rubric items whose annotations are confirmed correct divided by the total number of inspected rubric items. At least 90\% correctness is required to pass.

\paragraph{Annotation.}
Annotators enter the task after completing training and assessment. They annotate all \textbf{1,103 cases} and their rubrics, covering 328 clusters. Checks address medical plausibility, dialogue consistency, attribution to individuals and time points, and alignment of the current task with machine-generated rubrics; case content and criteria are corrected where needed. For Diffusion cases, annotators also check the medical meaning and task relevance of changes and the appropriate boundaries of uncertainty. Priorities are checked against the definitions in Section~\ref{sec:scoring}: P0 covers critical requirements concerning factual accuracy, core task completion, and medical safety; P1 covers information, explanations, or actions required for task completion; P2 covers additional value that does not alter the main conclusion.

\paragraph{Quality control and dispute resolution.}
After full annotation, 25\% of the data undergo quality inspection by a separate group with greater seniority and higher annotation accuracy. Inspectors verify consistency among inputs, tasks, and criteria, focusing on medical facts, scoring priorities, and within-cluster correspondences. If correctness is below 90\%, the full set is returned for reannotation; otherwise, identified errors are corrected before acceptance. Annotators may appeal findings through discussion within the annotation team. Unresolved disputes are escalated to chief-physician-level medical experts at tertiary Grade A hospitals.

\paragraph{Sampled acceptance.}
The final acceptance stage inspects a 10\% sample. At least 90\% correctness is required, followed by correction of identified errors before delivery; results below the threshold are returned for revision. Full annotation provides baseline coverage, while the two sampling stages and rework rules control delivery quality.

The resulting benchmark contains 1,103 cases and 4,931 rubric items. These correctness thresholds govern data production, whereas the 90.5\% agreement in Section~\ref{sec:human-judgment-agreement} concerns human verification of model-response judgments. The former checks cases and criteria; the latter checks whether the automated judge applies those criteria to responses consistently with human review.
\FloatBarrier

\section{Controlled Expansion and Clinical Examples (Diffusion)}
\label{sec:diffusion}

Diffusion constructs related evaluation tasks from a shared health setting, forming clusters whose members have equal status and different response requirements as individual facts or the current task change. Here, Diffusion refers to controlled case expansion, not a stochastic denoising generative model. Changes must be relevant to the health task and preserve a comparable background. Comparing these members examines whether models acquire currently valid information, integrate it with existing materials, and adjust or preserve their conclusions accordingly. In medical tasks, whether a laboratory value crosses a diagnostic threshold, which cause objective findings support, and changes in medication ingredients or doses may alter clinical judgments or management. If a corrected value does not cross the relevant threshold and the other applicable conditions remain unchanged, the response must use the corrected evidence while preserving the core medical conclusion.

\subsection{Medical Rationale: Conditional Judgments and Evidence Boundaries}
\label{sec:diffusion-medical-basis}
Clinical reasoning requires connecting current findings to an individual's background. Illness scripts organize relevant knowledge into enabling conditions, pathophysiological processes, and their manifestations; diagnostic experts likewise identify contextual conditions such as age and medical history as important components of disease knowledge organization~\citep{charlin2007scripts,custers1998illness}. The same complaint need not elicit the same response: differences in measurement protocols, values, pathogen evidence, and medication ingredients may alter the applicability of diagnostic criteria, action priorities, or the conditions for medication use.

From a probabilistic perspective, support for a disease hypothesis $d$ depends on evidence $e$ and individual context $c$:
\begin{equation}
 p(d\mid e,c)\propto p(e\mid d,c)\,p(d\mid c).
\end{equation}
We do not require models to output or calibrate posterior disease probabilities. Instead, we use this conditional structure to organize evaluation. Tasks with explicit diagnostic thresholds require identifying the measurement protocol and applicable population first; questions about causes or medications require distinguishing what the evidence supports from what remains unknown. Table~\ref{tab:clinical-axes} presents the expansion axes and medical constraints for the three examples below. Here, OGTT denotes the oral glucose tolerance test, and GDM denotes gestational diabetes mellitus.

\begin{table}[H]
\centering\hcltexttablefont
\caption{Medical judgment axes for controlled expansion. Within a shared task setting, evidence changes may require a different conclusion or preservation of a particular conclusion; both depend on the applicable conditions.}
\label{tab:clinical-axes}
\begin{tabularx}{\linewidth}{@{}L{29mm}L{39mm}Y@{}}
\toprule
\rowcolor{hcltexthead}
\textbf{Judgment axis} & \textbf{Conditions to control or distinguish} & \textbf{Implications for response requirements} \\
\midrule
Diagnostic thresholds and corrected values & The 75 g OGTT protocol at 24--28 weeks of gestation, measurement times, units, and corrected values & Reaching any corresponding threshold can change the GDM determination. If corrected values do not cross the relevant thresholds and the other applicable conditions remain unchanged, the core determination is preserved. Meeting diagnostic criteria and initiating insulin are separate questions~\citep{nhc2023gdm,ada2026pregnancy}. \\
\hcltextrowsep
Infection etiology and management & Parotid secretion cultures, identified pathogens, and the evidential status of serum IgM & Bacterial culture, identification, and antimicrobial susceptibility testing inform antibiotic treatment. Viral IgM provides supporting evidence rather than a definitive diagnosis on its own; management also considers transmission prevention and supportive care~\citep{cheng2026sialadenitis,cdc2024mumpsserology,cdc2025mumpsclinical}. \\
\hcltextrowsep
Medication ingredients and lactation contraindications & The complete formulation, specific product, and applicable local labeling & Avoid codeine-containing products during breastfeeding. Chinese labeling contraindicates oral single-ingredient dextromethorphan hydrobromide preparations during lactation; low milk exposure or absence of alcohol does not override this restriction~\citep{fda2017codeine,nmpa2021dextromethorphan}. \\
\bottomrule
\end{tabularx}
\end{table}

The medical rationale constrains both what should change and which limits on the conclusions should still apply. The same principle applies to hypoglycemia and breathing discomfort: a low glucose reading may require immediate management, but a single finger-stick measurement cannot establish the cause on its own~\citep{ada2026glycemic,cryer2009hypoglycemia}. A bronchodilator test can change the support for airflow limitation, while a single negative result cannot exclude asthma~\citep{gina2025strategy}. Appendix~\ref{sec:diffusion-examples} presents complete clusters for the three expansion axes above, and Appendix~\ref{sec:paired-responses} reports scores for the eight systems on those same members.

\subsection{From Task-Relevant Variables to Within-Cluster Variants}
Select a reference case within a cluster and denote it by $(C,q,R)$, where $q$ is the current question and its explicitly stated facts, $C$ is the remaining dialogue history and health materials, and $R$ is the set of rubric items. Evidence variables may occur in either $C$ or $q$. For expansions that change the evidence, we first identify a variable $d_k$ that may affect the task, then construct variants with different values or evidence states:
\begin{equation}
 d_k'\ne d_k,\qquad q_{\mathrm{task}}'=q_{\mathrm{task}}.
\end{equation}
Here, $q_{\mathrm{task}}$ denotes the semantics of the task requested by the question, not its verbatim wording. Task-relevant health facts remain part of the modified evidence even when they appear in the final question. Comparing a single variable requires coordinated updates to related statements; for example, correcting a laboratory field also requires updating the attribution of its numerical value. Other background information should remain consistent. Rewording irrelevant information can serve as an invariance control, whereas explicit corrections or state changes test the identification of currently valid evidence.

A second type of expansion changes the current question's objective, scope, or required response format. We interpret it as pure task redirection only when the complete context and current-turn evidence remain unchanged. If the question also introduces new individual circumstances, it is a task variant under changed conditions. The available evidence and rubrics must then be reassessed; the variant cannot simply be equated with switching tasks in the same setting.

The production process establishes the shared background, factor to change, and response objective in sequence, constructs the variant, generates matching rubrics, and checks consistency of facts, individuals, and time points. Medical review checks both whether the change itself is reasonable and whether the rubric items demand certainty beyond the evidence. This approach is consistent with behavioral testing and local contrast sets: it specifies what changed in the input and how the target behavior should respond, rather than merely checking whether two answers differ~\citep{ribeiro2020checklist,gardner2020contrast}. Such controlled comparisons operationalize health context learning through the identification and application of current evidence and satisfaction of response requirements.

Appendix~\ref{sec:diffusion-examples} presents complete construction examples involving diagnostic thresholds, pathogen evidence, and medication ingredients, including each member question, its shared identifier, and its scoring requirements.

\subsection{Cluster Composition and Evaluation Scope}
The benchmark contains 328 clusters comprising 1,103 cases, all included in both case-level and cluster-level evaluation. Each member has independent rubrics and equal evaluation status within its cluster; changing the reference member for presentation does not change the cluster score. Changes to evidence, states, or tasks may impose different response requirements. Evidence variations primarily examine whether responses track valid facts, whereas task variations examine whether a shared background supports different objectives; both may coexist within a cluster. Only comparisons with specific conditions controlled support an interpretation of a single changed factor, and not all within-cluster differences represent the same capability change.
\FloatBarrier

\section{Priority-Weighted Scoring with Geometric Cluster Aggregation}
\label{sec:scoring}

\bench{} evaluates responses using task-specific rubrics and aggregates results at both the case and related cluster levels. A case score measures how well an individual response satisfies the task requirements; a cluster score first treats related members from the same source as a unit and then summarizes their joint quality across clusters. All reported scores are converted to a 100-point scale. Full-context and query-only responses use the same score calculation; their differences describe performance changes under the reported conditions.

\subsection{Priorities and Judge Aggregation}
Each rubric specifies an independently assessable requirement, a polarity, and a priority. Positive criteria describe content that a response should satisfy, whereas negative criteria describe defects that should not occur. P0 denotes critical requirements concerning factual accuracy, core task completion, and medical safety; P1 denotes core information, explanations, or actions necessary to complete the task; and P2 denotes additional value that does not alter the main conclusion. The three priority weights are
\begin{equation}
 w(\mathrm{P0})=3,\qquad w(\mathrm{P1})=2,\qquad w(\mathrm{P2})=1.
\end{equation}
Semantically duplicate, subsuming, and positive--negative mirror requirements are merged to avoid assigning repeated weight to the same behavior. Each case has at most two P0 criteria and at least one valid criterion.

At the automatic judgment stage, Kimi K2.6, Qwen3.6-35B-A3B, and GLM-5.2 independently judge model responses against the self-contained rubrics. For each criterion, votes that cannot reach a judgment are excluded; an outcome is adopted when at least two judges agree, and is otherwise recorded as an unresolved judgment. The judgment concerns whether the content described by the rubric holds in the response: a positive criterion passes when it is met, and a negative criterion passes when it is not met. Judges do not infer criteria from task labels or treat the labels themselves as evidence of correctness. Section~\ref{sec:human-judgment-agreement} checks the agreement of these judgments with human review of sampled model responses.

\subsection{Case Scores with a Critical-Error Veto}
For criterion $j$ of case $i$, let $t_{ij}$ denote polarity, $\ell_{ij}$ priority, and $y_{ij}$ the adopted final judgment. The pass indicator is
\begin{equation}
 b_{ij}=\begin{cases}
 1,&t_{ij}=\mathrm{positive}\land y_{ij}=\mathrm{Met},\\
 1,&t_{ij}=\mathrm{negative}\land y_{ij}=\mathrm{Not\ Met},\\
 0,&\text{otherwise}.
 \end{cases}
 \label{eq:pass}
\end{equation}
The explicit failure indicator $f_{ij}$ equals 1 when a positive criterion is judged not met or a negative criterion is judged met; for an unresolved judgment, $b_{ij}=f_{ij}=0$. Let
\begin{align}
 W_i&=\sum_jw(\ell_{ij}),\qquad A_i=\sum_jw(\ell_{ij})b_{ij},\\
 D_i&=\sum_{j:\ell_{ij}\in\{\mathrm{P1},\mathrm{P2}\}}w(\ell_{ij})f_{ij},\\
 v_i&=\ind\!\left[\exists j:\ell_{ij}=\mathrm{P0}\land f_{ij}=1\right].
 \label{eq:veto}
\end{align}
The individual case score is defined as
\begin{equation}
 s_i=(1-v_i)\max\!\left(0,\frac{A_i-D_i}{W_i}\right)\in[0,1].
 \label{eq:case-score}
\end{equation}
An explicit failure on any P0 criterion gives the entire case a score of zero. In the absence of a P0 veto, passing P0/P1/P2 criteria contribute $+3/+2/+1$, respectively, while explicit failures on P1/P2 criteria contribute $-2/-1$, respectively, subject to a zero-score floor. Unresolved criteria add neither credit nor penalties, but their weights remain in the denominator. This rule distinguishes critical errors from noncritical deficiencies and prevents satisfaction of a subset of criteria from compensating for the omission of many necessary elements.

This score represents task utility defined by criterion priorities, rather than the probability that a response is correct. For a fixed $W_i$, in the absence of a P0 veto and before reaching the zero-score floor, changing one P1 criterion from pass to failure reduces the score by $4/W_i$, and the corresponding reduction for a P2 criterion is $2/W_i$: each change combines the loss of positive credit with an additional penalty. Fewer rubrics may therefore mean fewer constraints, but may also increase the relative impact of an individual error; the number of criteria alone cannot establish that one type of case is easier.

Scoring covers all predefined cases. In the implementation, $s_i$ is first rounded to four decimal places and then aggregated. For an evaluation set containing $N$ cases,
\begin{equation}
 S^{\mathrm{case}}=\frac{100}{N}\sum_{i=1}^{N}s_i.
 \label{eq:dataset-case}
\end{equation}

\subsection{Within-Cluster Geometric Means and Equal Weighting Across Clusters}
\label{sec:aggregate-operator}
An evaluation cluster consists of members connected by a shared setting and relationships defined by controlled changes. Let $\mathcal I_g$ be the set of member indices in cluster $g$, with $n_g=|\mathcal I_g|$. Its score is
\begin{equation}
 S_g=\left(\prod_{i\in\mathcal I_g}s_i\right)^{1/n_g},\qquad
 S^{\mathrm{cluster}}=\frac{100}{|\mathcal G|}\sum_{g\in\mathcal G}S_g.
 \label{eq:dataset-score}
\end{equation}
Here, $\mathcal G$ is the set of all 328 clusters. Each case is counted once within its cluster, and each cluster has equal weight in the dataset-level aggregate. If any member receives zero, the geometric cluster score is zero. The cluster-level metric thus emphasizes low-scoring members among related tasks and reduces the extent to which high-scoring cases compensate for local failures. It measures a different property from the arithmetic mean over all cases, so the numerical difference between the two cannot directly be interpreted as a change in model capability.

When $n_g=1$, the cluster score equals the member's case score. In a multimember cluster, a low-scoring member reduces the geometric score even when the other members score highly; a zero-scoring member makes the cluster score zero. Cluster evaluation therefore highlights weaknesses across related tasks. Section~\ref{sec:cluster-composition} examines performance by cluster size, and Appendix~\ref{sec:paired-responses} uses complete clusters to show how member scores jointly determine the cluster score.

The mean case score describes average satisfaction of individual task requirements, whereas the geometric cluster score emphasizes maintaining response quality across variations of a shared setting. Cluster scores are invariant to member order and the choice of reference question. Examining each member's response against its rubrics then identifies failures in interpreting evidence, recommending actions, or limiting conclusions to what the evidence supports.
\FloatBarrier

\section{Experimental Results and Analysis}
\label{sec:experiments}

We examine health context learning through three behavioral views supported by the resource. Full--Query comparisons describe task scores with and without supplementary background; case and complete-cluster results identify completion quality and weak members; subgroup statistics and item-level judgments connect differences to input requirements and unmet rubric criteria. Comparisons use the same benchmark and the scoring function defined in Section~\ref{sec:scoring}.

\subsection{Experimental Setup}
{\raggedright We compare GPT-6 Astra\citep{openai2026astra}, Claude Opus 5.5\citep{anthropic2026opus55}, Qwen3.8-27B\citep{alibaba2026qwenapi}, GLM-5.3\citep{zai2026glm53}, Kimi K3\citep{moonshot2026reasoning}, DeepSeek-V4.1-Flash\citep{deepseek2026models}, and Gemini 3.8 Flash\citep{google2026thinking}, and include \afuagent{}. The seven general-purpose models are configured to generate answers through official APIs in thinking mode with effort set to \texttt{high}, covering full-context and query-only conditions. \afuagent{} is an AI health assistant developed by Ant Group. We evaluate \afuagent{} under both full-context and query-only conditions.\par}

Full-context systems receive the health context and current question, without rubrics or target answers. Apart from the query-only comparison in Section~\ref{sec:context-ablation}, all main results, response analyses, and label profiles use full context. Both conditions for all eight systems use the benchmark's \textbf{1,103 cases and 328 clusters}, matched by case identity and evaluated with the scoring function in Section~\ref{sec:scoring}. Scores are displayed to two decimal places; differences use unrounded means. The analyses examine overall contextual answer quality, cross-condition score differences, performance across related tasks, and answering ability across context difficulties and medical categories in this task collection.

\paragraph{Human--judge agreement check.}
\label{sec:human-judgment-agreement}
To check the original automated rubric judgments, we randomly sampled 100 full-context responses from each model in the original seven-model evaluation, yielding 700 responses, and compared human and judge decisions item by item. Pooled across the rubric judgment items covered by these responses, agreement with the original judge was 90.5\%. Agreement is calculated as the number of rubric items with matching human and judge decisions divided by all reviewed items. This sampled check concerns the original automated decisions.

\subsection{Case-Level and Cluster-Level Performance with Full Context}
\label{sec:case-results}
\label{sec:main-results}
% Generated by analysis/revision_20261110_afu_query/render_assets.py.
\begin{table}[H]
\centering\hcldensetablefont
\caption{Overall results and context comparisons for all eight systems: 1,103 cases in 328 clusters.}
\label{tab:overall-results}
\label{tab:context-ablation-main}
\begin{tabularx}{\linewidth}{@{}L{24mm}|*{8}{>{\centering\arraybackslash}X}@{}}
\toprule
\rowcolor{hcltablehead}
Input / metric & \shortstack{GPT-6\\Astra} & \shortstack{Claude\\Opus 5.5} & \afuagent{} & \shortstack{Qwen3.8-\\27B} & GLM-5.3 & Kimi K3 & \shortstack{DeepSeek-\\V4.1-Flash} & \shortstack{Gemini\\3.8 Flash} \\
\midrule
\rowcolor{hcltablerow}
Full / Case & \textbf{75.03} & 68.21 & 61.99 & 57.31 & 52.29 & 48.46 & 45.71 & 42.77 \\ % key=Full.Case
Full / $G$ & \textbf{60.42} & 49.50 & 43.30 & 36.44 & 29.20 & 24.77 & 23.67 & 21.79 \\ % key=Full.$G$
\midrule
\rowcolor{hcltablerow}
Query / Case & \textbf{44.16} & 36.46 & 39.69 & 40.52 & 33.68 & 31.66 & 33.01 & 30.95 \\ % key=Query.Case
Query / $G$ & \textbf{21.45} & 14.78 & 17.75 & 17.03 & 12.33 & 9.74 & 12.43 & 10.92 \\ % key=Query.$G$
\midrule
\rowcolor{hcltablerow}
$\Delta$ / Case & +30.87 & +31.76 & +22.30 & +16.79 & +18.61 & +16.80 & +12.71 & +11.83 \\ % key=Delta.Case
$\Delta$ / $G$ & +38.97 & +34.72 & +25.56 & +19.41 & +16.87 & +15.03 & +11.24 & +10.87 \\ % key=Delta.$G$
\bottomrule
\end{tabularx}
\par\vspace{3pt}
\begin{minipage}{\linewidth}\fontsize{8.5}{9.6}\selectfont\raggedright Full: full context; Query: current question only; $\Delta=\mathrm{Full}-\mathrm{Query}$ (points). Case means weight cases equally; $G$ averages within-cluster geometric means across 328 equally weighted clusters. The seven general-purpose models use thinking/high; \afuagent{} is evaluated using Ant Group's AI health assistant service. Differences are descriptive comparisons. 
\end{minipage}
\end{table}

On the benchmark's full-context answering tasks, the eight systems have the same ordering on both metrics. GPT-6 Astra ranks first at 75.03 case points and 60.42 geometric cluster points, followed by Claude at 68.21/49.50, \afuagent{} at 61.99/43.30, and Qwen at 57.31/36.44. GLM, Kimi, DeepSeek, and Gemini follow in that order. GPT leads Claude by 6.82 case points and 10.92 cluster points; \afuagent{} leads Qwen by 4.68 and 6.86 points.

Individual-answer means and joint quality across related tasks still provide complementary information. GPT and Claude have 256 and 213 nonzero clusters, with mean geometric scores over their respective nonzero clusters of 77.42 and 76.23. Their overall cluster gap involves both nonzero-cluster coverage and completion quality. \afuagent{} has 194 nonzero clusters versus Qwen's 163, while their respective nonzero-cluster means are close, at 73.22 and 73.33. Across the eight systems, case scores range from 42.77 to 75.03 and cluster scores from 21.79 to 60.42.
Figure~\ref{fig:model-scores-grouped} provides the opening overview; Table~\ref{tab:overall-results} retains full system names, settings, and numerical results.

\subsection{Full Context versus the Current Question Alone}
\label{sec:context-ablation}

\paragraph{Paired design and measures.}
To examine the use of supplementary background in health context learning, we compare full-context ($F$) and query-only ($Q$) answers to the same current question, with and without that background. Query-only inputs retain the current question and its explicitly stated measurements, symptoms, and medication ingredients, while removing the other dialogue and health background. This design corresponds to the question-preserving comparison in CL-bench Life\citep{dou2026clbenchlife}. This section includes the seven general-purpose models and \afuagent{}, with settings described in Section~\ref{sec:experiments}. Both conditions cover the same 1,103 cases and 328 clusters, retaining identical current questions and rubrics for every case and using the penalty-based function in Section~\ref{sec:scoring}. The query-only inputs for \afuagent{} likewise remove historical messages, background embedded outside the current question, and other personal background information.

We define the Full--Query score difference as the difference in aggregate scores between the two input conditions on the same evaluation set:
\begin{equation}
 \Delta_{\mathrm{case}}=S_F^{\mathrm{case}}-S_Q^{\mathrm{case}},\qquad
 \Delta_{\mathrm{cluster}}=S_F^{\mathrm{cluster}}-S_Q^{\mathrm{cluster}}.
 \label{eq:context-gain}
\end{equation}
Positive values indicate higher task scores with full context. Cluster differences are calculated after geometric aggregation within complete clusters and equal weighting across clusters in each condition. Differences use unrounded means. This section compares the reported runs. The differences describe changes in scores between conditions and are not standalone estimates of causal effects. Full-context scores and Full--Query differences jointly characterize answer quality on complex questions; difficulty and scenario groups identify information acquisition, integration, and application tasks for follow-up evaluation.
Table~\ref{tab:context-ablation-main} consolidates the overall scores and differences for both input conditions.

\paragraph{Overall performance and coverage of related tasks.}
All eight systems score higher with full context, with case differences of 11.83--31.76 points and cluster differences of 10.87--38.97. GPT scores 44.16/21.45 with query-only inputs and 75.03/60.42 with full context, differences of 30.87 and 38.97. \afuagent{} scores 39.69/17.75 with query-only inputs and 61.99/43.30 with full context, differences of 22.30 and 25.56. Its query-only case score is slightly below Qwen's 40.52, whereas its cluster score is slightly above Qwen's 17.03; with full context, \afuagent{} exceeds Qwen on both metrics. Absolute scores and differences provide complementary views of task completion.

For GPT, 160 clusters score zero with query-only inputs and positively with full context; eight change in the reverse direction. The former transition occurs in 148, 125, 107, 101, 89, 84, and 75 clusters for Claude, \afuagent{}, Qwen, GLM, Kimi, DeepSeek, and Gemini, respectively. For \afuagent{}, 24 clusters change in the reverse direction, and nonzero-cluster coverage increases from 93 with query-only inputs to 194 with full context. Of GPT's 1,103 paired cases, 570 score higher, 478 are unchanged, and 55 score lower; the corresponding counts for \afuagent{} are 512, 449, and 142. The reported results also differ in nonzero-cluster coverage across related tasks.
% Generated by analysis/revision_20261110_afu_query/render_assets.py.
\usetikzlibrary{patterns,patterns.meta}
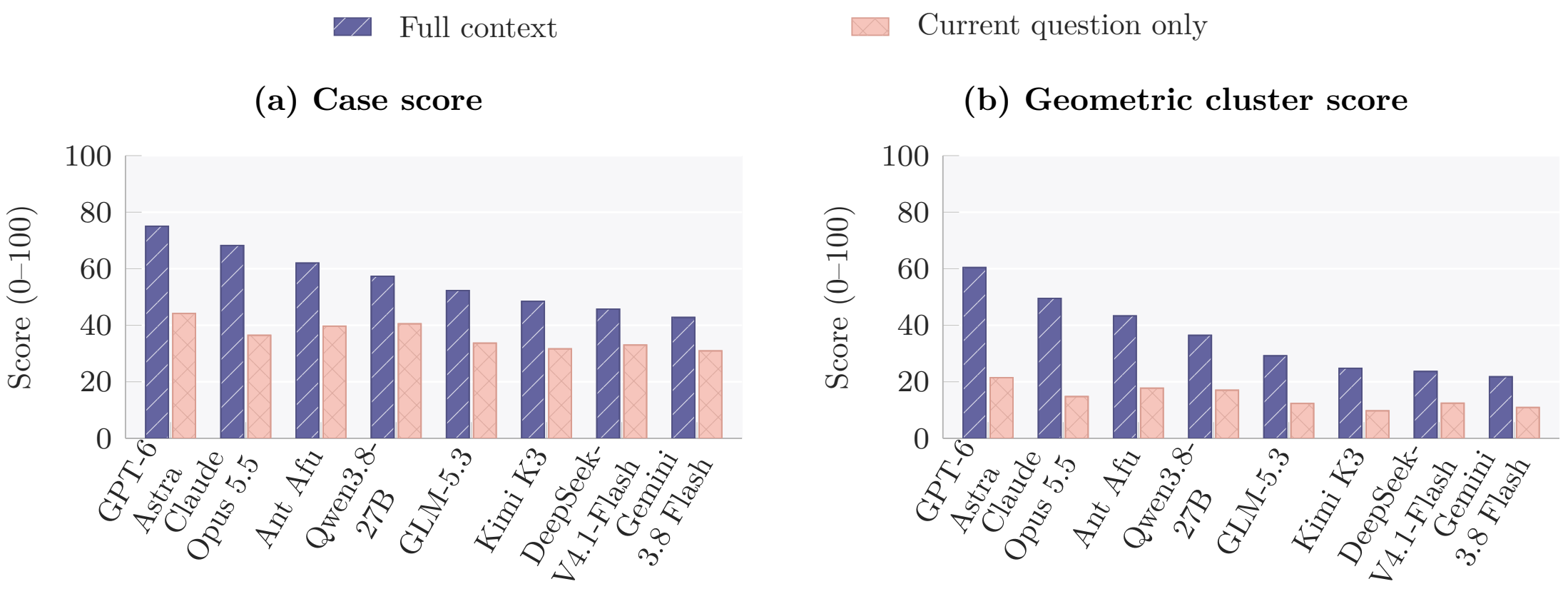
\begin{figure}[H]\centering
\begingroup
\definecolor{hclHatchPrimary}{HTML}{6464A0}
\definecolor{hclHatchPrimaryEdge}{HTML}{515183}
\definecolor{hclHatchSecondary}{HTML}{F6C5BC}
\definecolor{hclHatchSecondaryEdge}{HTML}{D99D91}
\definecolor{hclHatchBackground}{HTML}{F7F7F9}
\tikzset{
  hclHatchPrimary/.style={fill=hclHatchPrimary,draw=hclHatchPrimaryEdge,line width=0.35pt,
    postaction={pattern={Lines[angle=45,distance=4.5pt,line width=0.23pt]},pattern color=white!75!hclHatchPrimary}},
  hclHatchSecondary/.style={fill=hclHatchSecondary,draw=hclHatchSecondaryEdge,line width=0.35pt,
    postaction={pattern={Hatch[angle=45,distance=4.5pt,line width=0.23pt]},pattern color=hclHatchSecondaryEdge!85!white}}
}
\begin{tikzpicture}[x=1cm,y=1cm]
\draw[hclHatchPrimary] (0,0) rectangle (0.34,0.16);
\node[anchor=west,font=\fontsize{8}{9}\selectfont,text=black!85] at (0.48,0.08) {Full context};
\draw[hclHatchSecondary] (4.8,0) rectangle (5.14,0.16);
\node[anchor=west,font=\fontsize{8}{9}\selectfont,text=black!85] at (5.28,0.08) {Current question only};
\end{tikzpicture}\par\smallskip
\begin{minipage}[t]{0.445\linewidth}\centering
{\fontsize{8.5}{9.5}\selectfont\bfseries (a) Case score}\par\smallskip
\begin{tikzpicture}\begin{axis}[
width=\linewidth,height=42mm,ymin=0,ymax=100,ytick={0,20,40,60,80,100},
axis lines*=left,axis line style={draw=black!30,line width=0.35pt},tick style={draw=black!25},
tick label style={font=\fontsize{8}{9}\selectfont,text=black!85},
label style={font=\fontsize{8.2}{9.2}\selectfont,text=black!85},ylabel={Score (0--100)},
axis background/.style={fill=hclHatchBackground},ymajorgrids=true,grid style={draw=white,line width=0.5pt},
xmin=-0.6,xmax=7.6,xtick={0,1,2,3,4,5,6,7},xticklabels={{\shortstack{GPT-6\\Astra}},{\shortstack{Claude\\Opus 5.5}},{\afuagent{}},{\shortstack{Qwen3.8-\\27B}},{GLM-5.3},{Kimi K3},{\shortstack{DeepSeek-\\V4.1-Flash}},{\shortstack{Gemini\\3.8 Flash}}},
xticklabel style={rotate=60,anchor=east,font=\fontsize{8}{9}\selectfont},
bar width=6pt,clip=true]
\addplot[ybar,bar shift=0pt,forget plot,hclHatchPrimary] coordinates {(-0.18,75.030117860381)};
\addplot[ybar,bar shift=0pt,forget plot,hclHatchSecondary] coordinates {(0.18,44.160707162285)};
\addplot[ybar,bar shift=0pt,forget plot,hclHatchPrimary] coordinates {(0.82,68.212339075249)};
\addplot[ybar,bar shift=0pt,forget plot,hclHatchSecondary] coordinates {(1.18,36.456065276519)};
\addplot[ybar,bar shift=0pt,forget plot,hclHatchPrimary] coordinates {(1.82,61.990072529465)};
\addplot[ybar,bar shift=0pt,forget plot,hclHatchSecondary] coordinates {(2.18,39.685086128740)};
\addplot[ybar,bar shift=0pt,forget plot,hclHatchPrimary] coordinates {(2.82,57.305403445150)};
\addplot[ybar,bar shift=0pt,forget plot,hclHatchSecondary] coordinates {(3.18,40.517905711695)};
\addplot[ybar,bar shift=0pt,forget plot,hclHatchPrimary] coordinates {(3.82,52.288032638259)};
\addplot[ybar,bar shift=0pt,forget plot,hclHatchSecondary] coordinates {(4.18,33.677243880326)};
\addplot[ybar,bar shift=0pt,forget plot,hclHatchPrimary] coordinates {(4.82,48.457162284678)};
\addplot[ybar,bar shift=0pt,forget plot,hclHatchSecondary] coordinates {(5.18,31.655974614687)};
\addplot[ybar,bar shift=0pt,forget plot,hclHatchPrimary] coordinates {(5.82,45.714315503173)};
\addplot[ybar,bar shift=0pt,forget plot,hclHatchSecondary] coordinates {(6.18,33.008377153218)};
\addplot[ybar,bar shift=0pt,forget plot,hclHatchPrimary] coordinates {(6.82,42.774052583862)};
\addplot[ybar,bar shift=0pt,forget plot,hclHatchSecondary] coordinates {(7.18,30.945648232094)};
\end{axis}\end{tikzpicture}
\end{minipage}\hspace{0.01\linewidth}
\begin{minipage}[t]{0.445\linewidth}\centering
{\fontsize{8.5}{9.5}\selectfont\bfseries (b) Geometric cluster score}\par\smallskip
\begin{tikzpicture}\begin{axis}[
width=\linewidth,height=42mm,ymin=0,ymax=100,ytick={0,20,40,60,80,100},
axis lines*=left,axis line style={draw=black!30,line width=0.35pt},tick style={draw=black!25},
tick label style={font=\fontsize{8}{9}\selectfont,text=black!85},
label style={font=\fontsize{8.2}{9.2}\selectfont,text=black!85},ylabel={Score (0--100)},
axis background/.style={fill=hclHatchBackground},ymajorgrids=true,grid style={draw=white,line width=0.5pt},
xmin=-0.6,xmax=7.6,xtick={0,1,2,3,4,5,6,7},xticklabels={{\shortstack{GPT-6\\Astra}},{\shortstack{Claude\\Opus 5.5}},{\afuagent{}},{\shortstack{Qwen3.8-\\27B}},{GLM-5.3},{Kimi K3},{\shortstack{DeepSeek-\\V4.1-Flash}},{\shortstack{Gemini\\3.8 Flash}}},
xticklabel style={rotate=60,anchor=east,font=\fontsize{8}{9}\selectfont},
bar width=6pt,clip=true]
\addplot[ybar,bar shift=0pt,forget plot,hclHatchPrimary] coordinates {(-0.18,60.424731275358)};
\addplot[ybar,bar shift=0pt,forget plot,hclHatchSecondary] coordinates {(0.18,21.451023426758)};
\addplot[ybar,bar shift=0pt,forget plot,hclHatchPrimary] coordinates {(0.82,49.500240659763)};
\addplot[ybar,bar shift=0pt,forget plot,hclHatchSecondary] coordinates {(1.18,14.781597634661)};
\addplot[ybar,bar shift=0pt,forget plot,hclHatchPrimary] coordinates {(1.82,43.304163484191)};
\addplot[ybar,bar shift=0pt,forget plot,hclHatchSecondary] coordinates {(2.18,17.746861517558)};
\addplot[ybar,bar shift=0pt,forget plot,hclHatchPrimary] coordinates {(2.82,36.443405306297)};
\addplot[ybar,bar shift=0pt,forget plot,hclHatchSecondary] coordinates {(3.18,17.028588144073)};
\addplot[ybar,bar shift=0pt,forget plot,hclHatchPrimary] coordinates {(3.82,29.196484976580)};
\addplot[ybar,bar shift=0pt,forget plot,hclHatchSecondary] coordinates {(4.18,12.330402552650)};
\addplot[ybar,bar shift=0pt,forget plot,hclHatchPrimary] coordinates {(4.82,24.771133927628)};
\addplot[ybar,bar shift=0pt,forget plot,hclHatchSecondary] coordinates {(5.18,9.743522384691)};
\addplot[ybar,bar shift=0pt,forget plot,hclHatchPrimary] coordinates {(5.82,23.669110632779)};
\addplot[ybar,bar shift=0pt,forget plot,hclHatchSecondary] coordinates {(6.18,12.431731426735)};
\addplot[ybar,bar shift=0pt,forget plot,hclHatchPrimary] coordinates {(6.82,21.792379455611)};
\addplot[ybar,bar shift=0pt,forget plot,hclHatchSecondary] coordinates {(7.18,10.921796761562)};
\end{axis}\end{tikzpicture}
\end{minipage}
\endgroup
\caption{Full-context and current-question-only scores for all eight systems on 1,103 cases in 328 clusters. Both panels use a 0--100 vertical scale; diagonal hatching denotes Full and crosshatching Query. Differences are descriptive comparisons. }
\label{fig:context-ablation-scores}
\end{figure}

Query-only and full-context rankings differ. Claude's query-only case score is 36.46, below Qwen's 40.52, but Claude exceeds Qwen with full context, 68.21 versus 57.31. Their case differences are 31.76 and 16.79. At cluster level, Claude and Qwen score 14.78 and 17.03 with query-only inputs versus 49.50 and 36.44 with full context, differences of 34.72 and 19.41. Kimi's query-only case score of 31.66 is below DeepSeek's 33.01, but its difference of 16.80 exceeds DeepSeek's 12.71 and reverses their full-context case ranking. Joint reporting of absolute scores and differences preserves these distinct changes in performance.

\paragraph{Tasks and context structure.}
Table~\ref{tab:context-ablation-difficulty} in Appendix~\ref{sec:ablation-detail-results} groups differences by the complete task's scenario and primary difficulty, using the same paired cases in each group. GPT has its largest scenario difference in Health Management, at 33.17, followed by Clinical Decision-Making at 32.68. Claude's largest difference is in Care Triage, at 35.63, while Qwen's is in Health Management, at 17.87. GLM scores 24.44 higher in Medical Explanation. Cross-condition differences vary across models and answer objectives.

Across primary-difficulty groups, GPT has differences of 33.97, 33.88, and 33.65 on Complex, Conflicting, and Fragmented Context, respectively, well above its 11.28 on Other Difficulties. Claude's difference on Conflicting Context is 34.19, compared with Qwen's 18.43. \afuagent{} has its largest difficulty-group difference on Complex Context, at 25.67, versus 12.94 on Other Difficulties. In the reported evaluations, differences on Complex, Fragmented, and Conflicting Context exceed the same system's difference on Other Difficulties for all eight systems. Relation tracking, integration across fragments, and discrimination of currently valid information thus identify task types for targeted follow-up evaluation.

\paragraph{Dialogue turns.}
Full--Query differences do not increase monotonically with turn count. GPT scores 33.66 higher on 6--10 turns and 14.56 higher on 21--40 turns. In the latter group, \afuagent{}, Qwen, DeepSeek, and Gemini have differences of $-1.16$, $-11.97$, $-14.97$, and $-2.13$, respectively. Selecting relevant information, checking earlier interpretations, and integrating longer histories remain important follow-up targets. Single-turn cases can also contain reports or other health materials; GLM scores 26.42 higher on that group. The group above 40 turns has only 18 cases, where Claude's full-context case score is 82.56 versus GPT's 66.06, showing how performance on this group of long-context questions differs from the overall ranking. Appendix Table~\ref{tab:context-ablation-turns} reports each group's size and difference.

\subsection{Cluster Size and Weak Members}
\label{sec:cluster-composition}
% Generated by analysis/revision_20261110/render_assets.py.
\begin{table}[H]
\centering
\begin{minipage}[t]{0.49\linewidth}\vspace{0pt}% Generated by analysis/revision_20261110/render_assets.py.
\centering\hcldensetablefont
\caption{Geometric cluster means by cluster size.}
\label{tab:cluster-structure-size}
\begin{tabularx}{\linewidth}{@{}L{26mm}*{4}{>{\centering\arraybackslash}X}@{}}
\toprule
\rowcolor{hcltablehead}
System & \shortstack{1\\member\\$N=29$} & \shortstack{3\\members\\$N=135$} & \shortstack{4\\members\\$N=151$} & \shortstack{5\\members\\$N=13$} \\
\midrule
\rowcolor{hcltablerow}
GPT-6 Astra & 80.32 & \textbf{62.81} & \textbf{54.35} & \textbf{61.91} \\ % model=GPT-6-Astra
Claude Opus 5.5 & \textbf{84.41} & 50.19 & 43.07 & 39.19 \\ % model=Claude-Opus-5-5
\rowcolor{hcltablerow}
\afuagent{} & 75.33 & 44.31 & 35.95 & 46.84 \\ % model=Afu-Agent
Qwen3.8-27B & 74.50 & 40.60 & 25.82 & 31.81 \\ % model=Qwen3.8-27B
\rowcolor{hcltablerow}
GLM-5.3 & 58.73 & 30.92 & 23.22 & 14.82 \\ % model=GLM-5.3
Kimi K3 & 60.53 & 26.79 & 18.00 & 2.70 \\ % model=Kimi-K3
\rowcolor{hcltablerow}
\shortstack[l]{DeepSeek-\\V4.1-Flash} & 57.93 & 24.22 & 17.53 & 12.80 \\ % model=DeepSeek-V4.1-Flash
Gemini 3.8 Flash & 59.32 & 25.01 & 12.26 & 15.42 \\ % model=Gemini-3.8-Flash
\bottomrule
\end{tabularx}
\par\vspace{3pt}
\begin{minipage}{\linewidth}\fontsize{8.5}{9.6}\selectfont\raggedright Full context; 0--100 scale. $N$ counts clusters (328 total); clusters have equal within-group weights. Bold marks positive column maxima.
\end{minipage}
\end{minipage}\hfill
\begin{minipage}[t]{0.49\linewidth}\vspace{0pt}% Generated by analysis/revision_20261110/render_assets.py.
\centering\hcldensetablefont
\caption{Selected rubric checks in case cluster A.}
\label{tab:controlled-requirements}
\begin{tabularx}{\linewidth}{@{}L{26mm}*{4}{>{\centering\arraybackslash}X}@{}}
\toprule
\rowcolor{hcltablehead}
System & \shortstack{Changed\\A1$\to$A2} & \shortstack{Stable\\A1$\to$A3} & \shortstack{Update\\A3} & \shortstack{Compare\\A1$\to$A4} \\
\midrule
\rowcolor{hcltablerow}
GPT-6 Astra & 2/2 & 2/2 & 1/1 & 2/2 \\ % model=GPT-6-Astra
Claude Opus 5.5 & 2/2 & 2/2 & 1/1 & 2/2 \\ % model=Claude-Opus-5-5
\rowcolor{hcltablerow}
\afuagent{} & 1/2 & 0/2 & 1/1 & 1/2 \\ % model=Afu-Agent
Qwen3.8-27B & 2/2 & 2/2 & 1/1 & 2/2 \\ % model=Qwen3.8-27B
\rowcolor{hcltablerow}
GLM-5.3 & 2/2 & 2/2 & 1/1 & 2/2 \\ % model=GLM-5.3
Kimi K3 & 2/2 & 2/2 & 1/1 & 2/2 \\ % model=Kimi-K3
\rowcolor{hcltablerow}
\shortstack[l]{DeepSeek-\\V4.1-Flash} & 2/2 & 2/2 & 1/1 & 1/2 \\ % model=DeepSeek-V4.1-Flash
Gemini 3.8 Flash & 2/2 & 2/2 & 1/1 & 2/2 \\ % model=Gemini-3.8-Flash
\bottomrule
\end{tabularx}
\par\vspace{3pt}
\begin{minipage}{\linewidth}\fontsize{8.5}{9.6}\selectfont\raggedright Satisfied/selected counts. The source criteria are A1:1+A2:1, A1:1+A3:1, A3:2, and A1:1+A4:3, respectively; columns must not be summed.
\end{minipage}
\end{minipage}
\par\vspace{3pt}
\begin{minipage}{\linewidth}\fontsize{8.5}{9.6}\selectfont\raggedright The seven general-purpose models use thinking/high; \afuagent{} is evaluated using Ant Group's AI health assistant service. The right panel matches selected shared requirements to the existing final judgments of complete rubric items.
\end{minipage}
\end{table}

The following analysis partitions clusters into disjoint subgroups by member count and reports mean cluster scores within each subgroup.

In the subgroup of 29 singleton clusters, Claude scores 84.41, above GPT's 80.32; \afuagent{} and Qwen score 75.33 and 74.50. In the subgroup of 135 three-member clusters, GPT leads at 62.81, followed by Claude, \afuagent{}, and Qwen at 50.19, 44.31, and 40.60. The subgroup of 151 four-member clusters, covering 604 cases, is the largest; these four systems score 54.35, 43.07, 35.95, and 25.82, respectively. GPT's lead on multi-member clusters differs from the singleton ordering.

In the subgroup of 13 five-member clusters, GPT, \afuagent{}, Claude, and Qwen score 61.91, 46.84, 39.19, and 31.81, respectively. GPT, \afuagent{}, and Qwen have higher scores in the five-member subgroup than in the four-member subgroup, whereas Claude has a lower score. These subgroups differ in task composition, and their scores do not decline monotonically with member count. \afuagent{} also exceeds Claude on five-member clusters, showing why aggregate rankings cannot replace examination of specific task structures.

\paragraph{Zero-scoring cases and critical errors.}
% Updated by analysis/revision_20261110_zero_presentation/update_presentation.py.
\begin{figure}[H]\centering
\usetikzlibrary{patterns,patterns.meta}
\begingroup
\definecolor{hclHatchPrimary}{HTML}{6464A0}
\definecolor{hclHatchPrimaryEdge}{HTML}{515183}
\definecolor{hclHatchSecondary}{HTML}{F6C5BC}
\definecolor{hclHatchSecondaryEdge}{HTML}{D99D91}
\definecolor{hclHatchBackground}{HTML}{F7F7F9}
\tikzset{
  hclHatchPrimary/.style={fill=hclHatchPrimary,draw=hclHatchPrimaryEdge,line width=0.35pt,
    postaction={pattern={Lines[angle=45,distance=4.5pt,line width=0.23pt]},pattern color=white!75!hclHatchPrimary}},
  hclHatchSecondary/.style={fill=hclHatchSecondary,draw=hclHatchSecondaryEdge,line width=0.35pt,
    postaction={pattern={Hatch[angle=45,distance=4.5pt,line width=0.23pt]},pattern color=hclHatchSecondaryEdge!85!white}}
}

\begin{tikzpicture}\begin{axis}[
width=0.86\linewidth,height=54mm,xbar stacked,bar width=9pt,
axis lines*=left,axis line style={draw=black!30,line width=0.35pt},xmin=0,xmax=400,ymin=-0.6,ymax=7.6,y dir=reverse,
ytick={0,1,2,3,4,5,6,7},yticklabels={{GPT-6 Astra},{Claude Opus 5.5},{\afuagent{}},{Qwen3.8-27B},{GLM-5.3},{Kimi K3},{DeepSeek-V4.1-Flash},{Gemini 3.8 Flash}},
xlabel={Number of zero-score cases},tick label style={font=\fontsize{8.4}{9.3}\selectfont,text=black!85},
label style={font=\fontsize{8.2}{9.2}\selectfont,text=black!85},tick style={draw=black!25},axis background/.style={fill=hclHatchBackground},xmajorgrids=true,grid style={draw=white,line width=0.5pt},
legend style={font=\fontsize{8}{9}\selectfont,draw=none,fill=white,at={(0.5,1.02)},anchor=south,column sep=8pt},legend columns=2]
\addplot[hclHatchPrimary,point meta=explicit symbolic,nodes near coords,nodes near coords align={horizontal},every node near coord/.append style={font=\fontsize{8}{9}\selectfont,text=black!90,anchor=east,fill=white,fill opacity=0.94,text opacity=1,inner xsep=0.8pt,inner ysep=0.3pt,xshift=-2pt}] coordinates {(58,0)[58] (81,1)[81] (107,2)[107] (124,3)[124] (128,4)[128] (150,5)[150] (170,6)[170] (152,7)[152]};
\addplot[hclHatchSecondary,point meta=explicit symbolic,nodes near coords,nodes near coords align={horizontal},every node near coord/.append style={font=\fontsize{8}{9}\selectfont,text=black!90,anchor=east,fill=white,fill opacity=0.94,text opacity=1,inner xsep=0.8pt,inner ysep=0.3pt,xshift=-2pt}] coordinates {(27,0)[27] (67,1)[67] (85,2)[85] (124,3)[124] (136,4)[136] (160,5)[160] (165,6)[165] (192,7)[192]};
\legend{P0 veto,Non-P0 zeros}
\end{axis}\end{tikzpicture}\endgroup
\caption{Zero-scoring cases under full context, grouped by whether a P0 veto is triggered. The two categories are mutually exclusive and cover all zero-scoring cases.}
\label{fig:cluster-structure-zero-modes}
\end{figure}
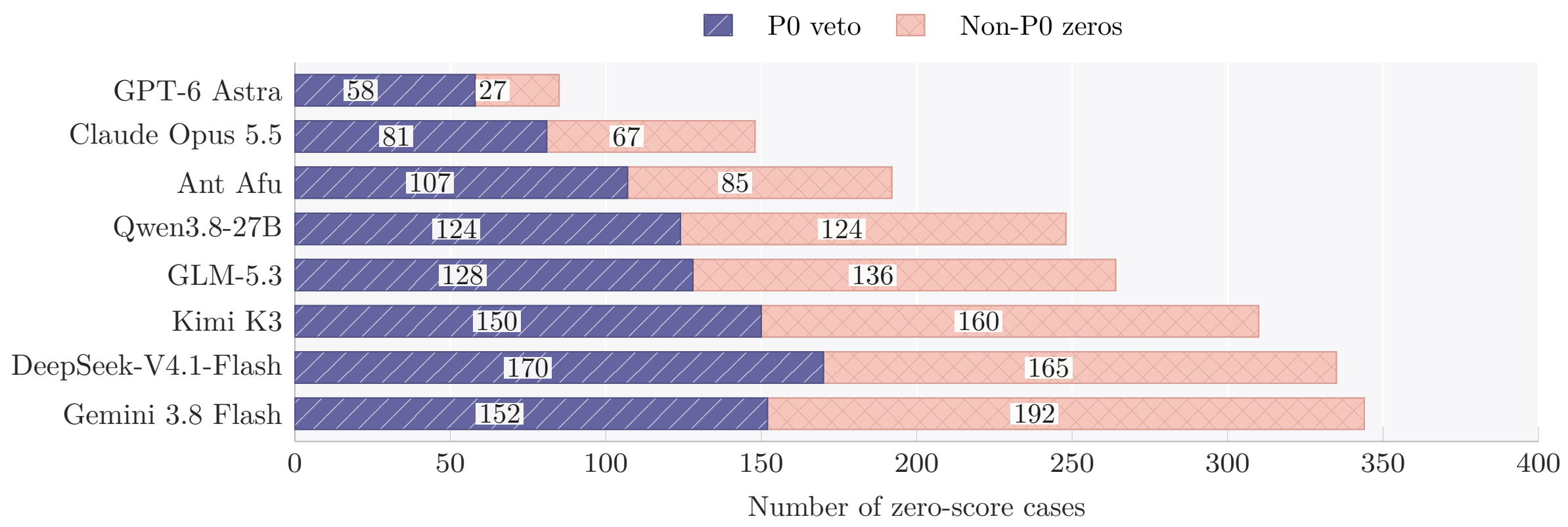
Under full context, zero-scoring cases are grouped into those with and without a P0 veto. The corresponding counts are 58 and 27 for GPT, 81 and 67 for Claude, 107 and 85 for \afuagent{}, and 124 and 124 for Qwen. P0 vetoes highlight critical facts or safety requirements; non-P0 zeros comprise the remaining zero-scoring results. These counts characterize the composition of zero scores across systems, while specific failures require examination of the responses and rubrics. Section~\ref{sec:evidence-boundaries} further relates the GPT--Claude difference to scoring mechanisms and actual responses.

\subsection{Responses and Judgments within Complete Related Clusters}
\paragraph{A three-member comparison.}
\label{sec:paired-main}
The three-member example in Table~\ref{tab:cluster-walkthrough} connects context changes to checkable requirements; Table~\ref{tab:case-b-scores} reports all eight systems' scores. GPT-6 Astra scores 100 on all three members. Claude Opus 5.5, Kimi K3, Qwen3.8-27B, and DeepSeek-V4.1-Flash obtain geometric cluster scores $G$ of 78.09, 69.33, 67.40, and 65.87, respectively. \afuagent{}, GLM-5.3, and Gemini 3.8 Flash each have a zero-scoring member, giving $G=0$. Claude identifies acute suppurative parotitis in B1 and makes susceptibility-guided antibacterial treatment the main management approach, passing both P0 criteria under the rubric judgments. The P1 criterion requiring post-aspiration follow-up to assess reaccumulation and further drainage when needed is unmet, giving 71.43. GLM satisfies B3's P0 requirement to prioritize dentistry, but both P1 criteria---ENT assessment when mainly throat or palate pain remains, and earlier care upon deterioration before Friday---are unmet, flooring its net score at zero. These records link cluster scores to members, criteria, and response evidence for item-level inspection. Appendix~\ref{sec:paired-responses} gives every member score for three example clusters, paired with the questions and summarized rubrics in Appendix~\ref{sec:diffusion-examples}.

\paragraph{Updating and preserving requirements under controlled changes.}
Turning to the four-member OGTT cluster (Cluster A), Table~\ref{tab:controlled-requirements} examines selected rubric criteria. The four members share the same dialogue history: A1's 2-hour value is 8.31 mmol/L; A2 raises it to 8.81, requiring a changed conclusion; A3 lowers it to 8.21, requiring an updated value and the same core conclusion. A4 retains the original values and shifts to insulin consultation, while the threshold comparison must remain. The seven general-purpose models satisfy the selected requirements in the first three columns. \afuagent{} uses A3's corrected value but fails the pregnancy-threshold determination required in A1 and A3, giving counts of 1/2, 0/2, 1/1, and 1/2 across the four columns. In the last column, DeepSeek-V4.1-Flash also fails the selected A4 criterion: its answer labels 8.31 mmol/L as slightly above 8.5, and all three judges mark it as unmet. This example distinguishes using an updated value, reaching the corresponding determination, and retaining an applicable comparison after the task changes.

\paragraph{Two complete four-member clusters.}
To further connect member scores with cluster scores, Table~\ref{tab:cluster-value-examples} presents all four members of the breathing-discomfort and diet-and-sleep clusters. The breathing cluster shares a history of insufficient breath when singing, hoarseness, occasional reflux, and earlier cardiac-test findings. Its members are the reference question, variants adding positive and negative bronchodilator-test results, and a task requesting cardiac, pulmonary, and laryngeal assessment plans. The diet cluster shares recent weight fluctuations and insufficient sleep. Its members concern early awakening after reduced food intake, fingertip glucose readings at awakening of 2.9 and 5.7 mmol/L, and whether the week's weight changes establish fat loss. Each case is scored against its own rubrics; $G$ is calculated from case scores rounded to four decimal places on $[0,1]$ and converted to the 100-point scale. These selected clusters illustrate scoring paths rather than estimate population error frequencies. Quotations below are translated from the evaluated Chinese answers.

% Generated by analysis/revision_20261110/render_assets.py.
\begin{table}[H]
\centering\hcldensetablefont
\caption{All member scores and geometric cluster scores $G$ for two complete example clusters (full context; 0--100 scale).}
\label{tab:cluster-value-examples}
\begin{tabularx}{\linewidth}{@{}L{35mm}*{5}{>{\centering\arraybackslash}X}@{}}
\toprule
\rowcolor{hcltablehead}
System & Reference & Variant 1 & Variant 2 & Task variant & $G$ \\
\midrule
\rowcolor{hcltableband}
\multicolumn{6}{@{}l@{}}{\textbf{Breathing discomfort: symptom interpretation and test planning}} \\
GPT-6 Astra & 73.33 & 100.00 & 60.00 & 100.00 & 81.44 \\
Claude Opus 5.5 & 73.33 & 100.00 & 100.00 & 100.00 & 92.54 \\
Qwen3.8-27B & 33.33 & 0.00 & 60.00 & 83.33 & 0.00 \\
\rowcolor{hcltableband}
\multicolumn{6}{@{}l@{}}{\textbf{Diet and sleep: glucose evidence and fat-loss assessment}} \\
GPT-6 Astra & 100.00 & 100.00 & 55.56 & 100.00 & 86.34 \\
Claude Opus 5.5 & 100.00 & 63.64 & 55.56 & 100.00 & 77.11 \\
\bottomrule
\end{tabularx}
\par\vspace{3pt}
\begin{minipage}{\linewidth}\fontsize{8.5}{9.6}\selectfont\raggedright $G$ uses all four members.
\end{minipage}
\end{table}

\paragraph{Breathing: assessment planning and evidence-variant failures can coexist.}
The current Qwen run's planning answer lists electrocardiography, pulmonary-function testing with bronchodilator testing, and electronic laryngoscopy, passing the three-system P0 criterion. Only the P2 criterion on cardiopulmonary exercise testing is unmet, giving 83.33. Its positive-test member states that ``your asthma is now confirmed'' and refers to ``chest tightness, worse when lying down,'' whereas the history records no worsening when lying down. Three P1 criteria on diagnostic qualification, the cardiac-judgment boundary, and the positional fact are unmet. Penalties floor the net score at zero, giving $G=0$ for the cluster. GPT scores 100 on the positive-test and planning members, but misses the reference question's P1 criterion on explicit respiratory or cardiology referral and the negative-test member's compound P1 criterion on ENT assessment despite mentioning laryngeal factors. These members score 73.33 and 60.00, giving $G=81.44$. Claude's positive-test member retains the qualification that ``the doctor who ordered the test makes the final diagnosis.'' Its negative-test and planning members also meet every criterion. The reference question misses the P1 criterion on a complete set of warning signs and scores 73.33; the other three score 100, giving $G=92.54$. Claude's higher joint score here shows why the overall ranking cannot replace inspection of each evidence branch.

\paragraph{Diet: identical member scores can follow different unmet criteria.}
GPT and Claude both score 100 on the reference question and the fat-loss task. On the 2.9 mmol/L member, both pass the P0 criterion for prioritizing hypoglycemia management and repeat measurement, but Claude incompletely states the restriction on oral intake when consciousness or safe swallowing is impaired. That P1 criterion is unmet, giving 63.64. Both score 55.56 on the 5.7 mmol/L member: GPT misses the P1 criterion on targeted recovery advice tied to the recorded substantial sleep deficit, while Claude misses the P1 criterion on medical assessment for persistent or recurrent sleep problems. Across all four members, GPT and Claude obtain $G=86.34$ and $77.11$. The complete records separate satisfaction of immediate management, judgment boundaries, and follow-up actions; equal scores, individual perfect scores, and cluster scores alone do not identify the specific omission.

\subsection{Evidence Use and Task Requirements in Contextual Answering: GPT versus Claude}
\label{sec:evidence-boundaries}
GPT exceeds Claude by 6.82 case points and 10.92 geometric cluster points. Using the rubric judgments adopted in this paper, we connect the difference to specific requirements for \emph{faithfulness to evidence, appropriately qualified conclusions, and complete action recommendations} in complex health question answering. Models must distinguish user-provided facts from earlier responses' inferences, match conclusions to evidence strength, and make subsequent recommendations conditional on any unresolved prerequisites. Medical evaluation's emphasis on scientific consensus, potential harm, and completeness motivates these requirements\citep{singhal2023clinical}.

\paragraph{Where the score difference arises in this benchmark.}
Both models have valid judgments for all 1,103 cases. Matching the same criterion set, GPT and Claude pass 94.16\% and 91.73\% of 1,028 P0 criteria, and 84.13\% and 79.19\% of 3,648 P1 criteria; unresolved criteria remain in the denominator. On 255 P2 criteria, Claude passes 69.41\%, above GPT's 64.31\%, so the direction is not uniform across priorities. Decomposing the overall case-score difference, one-sided P0 vetoes contribute a net 1.77 points, while zero net scores after non-P0 penalties contribute 2.28 points. Together they account for 4.04 points, or 59.30\% of the total gap. Each contribution sums paired score differences for cases in which one response scores above zero and the other scores zero for the corresponding reason, then divides by 1,103, including reverse-direction differences as losses; totals use unrounded values. The remaining 2.77 points come from 908 cases with two positive scores. This describes score composition under the rubric judgments, rather than treating zero counts as clinical-error incidence.

GPT scores 100.00 on each of the following three selected cases, while Claude scores 0.00, 0.00, and 33.33. They illustrate correction of earlier inferences, limits of causal judgments, and differences in the basis for management. They explain specific criteria rather than represent the direction of every task difference; the complete breathing cluster in Section~\ref{sec:paired-main} also illustrates a higher cluster score for Claude. Quotations are translated from the evaluated Chinese answers.

\Needspace{7\baselineskip}
\paragraph{Red particles in a newborn's stool: verify test evidence and temporal relationships.}
A parent reports red particles in a 27-day-old infant's stool, describes a stool examination as unremarkable, and adds that the vitamin A and D supplement is a clear oil and lactase is a white powder started yesterday. GPT states that ``a normal stool examination does not mean bleeding or infection has been excluded'' and asks, ``Did the red particles appear before or after lactase was started?'' It treats appearance and the reported test result as clues requiring verification. Claude also corrects the earlier explanation that capsule material could pass through breast milk into the infant's stool. However, after proposing urate staining and surface discoloration as possibilities, it says these are ``not bleeding'' and relies on ``a normal occult-blood result in the stool examination,'' which the user did not provide. Both give follow-up care advice, but GPT retains the boundary that bleeding has not been excluded, while Claude's exclusion exceeds the available evidence. The rubric judgments for the P0 uncertainty criterion are pass and explicit failure, producing scores of 100.00 and 0.00.

\paragraph{Medication-associated tinnitus: distinguish a suspected trigger from diagnostic confirmation.}
The user reports tinnitus approximately half an hour after each diclofenac dose and its absence without the drug, a pattern lasting nearly a month; the history also mentions affected hearing. Both models identify diclofenac as the leading suspected trigger and recommend prescription review and otolaryngology assessment. GPT states that ``whether hearing damage has occurred still requires audiometric confirmation'' and that ``even if tinnitus resolves after stopping the drug, otoscopy and pure-tone audiometry are still recommended.'' Claude, conditional on tinnitus resolving and hearing recovering within days of discontinuation, says that ``this can essentially confirm that the drug caused it.'' Its two P1 criteria on investigation and prescription advice pass, but the P0 criterion limiting the diagnostic force of the temporal relationship is judged unmet. Claude therefore scores 0.00 and GPT 100.00. Identifying a valuable causal clue still requires separating grounds for suspicion from the evidence needed for confirmation.

\paragraph{Conflicting eyelid diagnoses: make evidence priority govern management.}
An in-person assessment diagnosed a stye, while the user describes diffuse redness, swelling, and marked itching. Both models discuss eyelid dermatitis or eczema. Conditional on this having been an in-person ophthalmologist's assessment, GPT says to ``use the in-person prescription as the basis, rather than switching to an eczema regimen solely from an online description,'' and requests the actual prescription. Claude acknowledges that online assessment cannot replace examination, asks about medication, and recommends a return visit. Yet its symptom comparison concludes that ``eyelid eczema (allergic) currently appears more likely,'' without explicitly organizing current management around the in-person diagnosis. The P1 criterion on this priority is judged unmet, while the other two criteria pass, producing 33.33; GPT meets all three and scores 100.00. The distinction concerns consistency among diagnostic preference, evidence hierarchy, and subsequent action, beyond merely naming a possible condition.

\paragraph{Relation to prior medical evaluations.}
HealthBench's consensus-criterion analysis directly compares GPT-4.1 with Claude 3.7 Sonnet (extended thinking). Their pass rates for appropriate qualification under reducible uncertainty ($N=285$) are 98.25\% and 89.82\%; under irreducible uncertainty ($N=198$), 91.92\% and 82.83\%; and for direct answers when the available information is sufficient ($N=228$), 96.93\% and 93.42\%\citep{arora2025healthbench}. These results for the specified model versions illustrate separate criteria for qualified judgments and sufficiently supported answers, informing the dimensions used here to examine contextual evidence use.

An independent study also found that similar overall clinical ratings can coexist with a larger gap on fine-grained requirements. Comparing GPT-5.2 and Claude Opus 4.6, Vishwanath et al. reported blinded clinician ratings of 3.54 and 3.52 on a four-point scale for non-refusal responses in a study of 100 real clinical queries, but multimodel-judged scores of 88.0 and 77.0 on 500 HealthBench items\citep{vishwanath2026clinicalcomparison}. Overall human ratings and fine-grained criteria provide different views of responses in that study. These studies motivate examining uncertainty and completeness separately, complementing the criterion-level analysis above.

The criterion rates and response comparisons locate some differences in this benchmark in checking earlier information, matching conclusions to evidence, and making recommendations that respect the relevant conditions. The external studies further motivate these evaluation dimensions, but involve different model versions and judging conditions and do not explain the cause of the present score gap. P0 vetoes and P1/P2 penalties express criterion differences at case level, while geometric aggregation presents joint quality across related members. GPT and Claude have 72 and 115 zero-scoring clusters, respectively, yet Claude scores higher on the complete breathing cluster, and P2 pass rates also favor Claude. Overall ranking, local strengths, and specific scoring mechanisms therefore require joint inspection. These results describe the recorded responses, run settings, and rubric judgments without isolating their respective effects on the gap.

\subsection{Profiles Across Tasks and Context Difficulties}
\label{sec:label-results}
Detailed numerical results are provided in Appendix~\ref{sec:full-category-results}; this section uses the main figures and representative differences to explain the contextual capability profile.
Group results characterize contextual answering under different task conditions and are equally weighted case means under full context. Medical scenarios distinguish response objectives, primary context difficulties distinguish information-processing demands, and disease and department labels describe medical content. The first two form separate mutually exclusive partitions, while the latter two can overlap. Each group uses the same cases and scoring function for all eight systems, and its scores reflect its specific contexts and rubric requirements.

\input{figures/capability_profiles}

\paragraph{Medical scenarios.}
Figure~\ref{fig:capability-profiles}(a) shows scores by medical scenario.
GPT leads all five scenarios, ranging from 72.50 to 81.84. All eight systems achieve their highest scenario score in Care Triage, performing comparatively well on this group under the scoring scheme. Health Assessment and Clinical Decision-Making require linking individual facts to judgments and actions and jointly test information integration and task application in these questions.

Claude scores 68.83, 80.38, and 66.48 in Health Assessment, Care Triage, and Health Management; \afuagent{} scores 61.81, 70.88, and 58.49, and Qwen scores 57.45, 66.96, and 52.93. In Medical Explanation, \afuagent{} slightly exceeds Claude, 59.84 versus 58.84, while GLM's 55.48 exceeds Qwen's 51.41, revealing local strengths that differ from the overall ranking. This group contains only 41 cases, with its coverage reported alongside the scores.

\paragraph{Context difficulties.}
Figure~\ref{fig:capability-profiles}(b) shows scores by primary context difficulty.
GPT leads all primary-difficulty groups, scoring 75.82 on Conflicting Context and 72.19 on Fragmented Context. Claude scores 66.24 and 66.38 on these groups, \afuagent{} scores 55.78 and 54.86, and Qwen scores 53.06 and 54.40. On Noisy Context, Claude, \afuagent{}, and Qwen score 77.60, 72.83, and 67.66. Kimi exceeds DeepSeek on Conflicting Context, 45.84 versus 41.54, while DeepSeek exceeds Kimi on Complex Context, 47.06 versus 46.25. These groups impose different contextual demands, including relation tracking, evidence integration, and assessment of conflicting information, revealing answering differences beyond the overall score.

\paragraph{Diseases and departments.}
GPT leads seven of the eight main departments and seven of the eight main disease categories shown. Claude leads Endocrinology at 74.54 versus GPT's 73.61 and Diabetes at 73.17 versus GPT's 71.81. Claude exceeds Qwen across these main department and disease groups; examples include Gastroenterology at 68.52 versus 56.40 and Fungal Infection at 74.98 versus 63.74. \afuagent{} scores 72.66 on Fungal Infection and 58.47 on Anemia, between Claude and Qwen in both groups. These differences characterize contextual answering on tasks involving different medical content.

Middle-ranked models are also complementary. Kimi scores 50.67 on Health Management, above DeepSeek's 41.22. DeepSeek instead scores 46.74 on Clinical Decision-Making and 44.44 on Medical Explanation, slightly above Kimi's 45.16 and 43.40. Combining task, difficulty, and domain analyses locates overall advantages in concrete information-use and answer requirements, beyond a single ranking.
\FloatBarrier

\section{Related Work}
\label{sec:related}

\paragraph{Medical question answering and specific response requirements.}
MedQA and MedMCQA assess knowledge application and reasoning through medical multiple-choice examinations~\citep{jin2021medqa,pal2022medmcqa}. MultiMedQA covers examinations, medical research, and consumer health questions, and the Med-PaLM series further assesses open-ended response quality through human evaluation~\citep{singhal2023clinical,singhal2025medpalm2}. HealthBench uses physician-authored, conversation-specific rubrics to evaluate the final response to a health conversation. Its interaction design includes ambiguous expressions, information supplied across turns, and errors in earlier answers, and its evaluation includes context awareness~\citep{arora2025healthbench}. These designs establish foundations for evaluating health answers in context.

\paragraph{Chinese-language clinical and health evaluation.}
CMB-Clin contains 74 complex cases and 208 questions, organizing case descriptions and preceding questions and answers into multi-turn diagnostic and treatment tasks. Responses are assessed for fluency, relevance, completeness, and medical proficiency~\citep{wang2024cmb}. ClinConsensus v6 contains 2,500 Chinese open-ended cases with 30 binary rubric criteria per case and physician-calibrated coverage thresholds~\citep{zheng2026clinconsensus}. CliMedBench covers 14 clinical scenarios using materials including real patient records, guidelines, literature, and online consultations~\citep{ouyang2024climedbench}. MedBench v4 by Ding et al. provides text, multimodal, and agent evaluation; MedLongConv and MedLongQA examine long-conversation memory and information extraction from extended clinical documents~\citep{ding2025medbenchv4}. Chinese-language tasks, multi-turn interaction, authentic medical materials, and open-ended criteria thus have established precedents. Building on these precedents, our resource jointly designs open-ended tasks, explicit context-difficulty labels, controlled case variations within shared health settings, and member-specific medical evaluation criteria, so that changes in evidence, patient states, or task goals have corresponding response requirements.

\paragraph{Related examples and changes in medical conditions.}
CheckList's invariance and directional tests and Contrast Sets' meaningful local modifications provide foundations for evaluating related examples~\citep{ribeiro2020checklist,gardner2020contrast}. In medicine, MediEval jointly assesses statements' medical correctness and support from patient records~\citep{qu2026medieval}. MedEinst changes discriminating evidence in paired cases and evaluates diagnostic inertia through labels and pair-level metrics~\citep{chen2026medeinst}. MedPIC-Bench changes patient information affecting medication-rule applicability and evaluates responses through option exact match and pair accuracy~\citep{hou2026medpic}. \bench{} extends such relational evaluation to open-ended requirements in explanation, assessment, decision-making, triage, and health management. Member variations may require revising a judgment, retaining a conclusion after updating evidence, or preserving still-applicable facts and action prerequisites under a new task.

\paragraph{Acquiring and applying contextual information.}
LongBench, Lost in the Middle, and RULER examine long-input use through task coverage, information position, and controllable complexity~\citep{bai2024longbench,liu2024lost,hsieh2024ruler}. In CL-bench, domain experts construct multiple tasks and task-level rubrics under shared contexts, emphasizing the application of new contextual knowledge~\citep{dou2026clbench}. CL-bench Life extends evaluation to everyday communication, fragmented materials, and behavioral records, including health records~\citep{dou2026clbenchlife}. ASCENT examines clinical diagnostic impressions and their updates as evidence is progressively provided~\citep{choi2026ascent}. We instantiate these information-processing demands through subject and time attribution, expression understanding, evidence integration, and assessment of conflicting information in Chinese-language health tasks. Full--Query comparisons describe score differences between inputs with and without supplementary background, while clusters and rubrics locate unmet requirements across related tasks. Medical annotation, quality control, and acceptance checks support alignment among inputs, member relations, and scoring criteria.

\section{Discussion and Limitations}
\label{sec:discussion}

\paragraph{Targeted follow-up evaluation.}
Full--Query comparisons can show score differences with and without specified background. Selecting tasks by scenario and difficulty can form regression sets focused on relation tracking, cross-turn evidence integration, or assessment of conflicting information for comparing model versions or training approaches. Such evaluations should retain the original inputs, rubrics, and scoring rules, keep judgment policies consistent, and examine both individual cases and complete clusters, avoiding selection of only a well-performing member.

The complete clusters in Appendix~\ref{sec:paired-responses} reveal local strengths and shared weaknesses. Claude and GLM satisfy all requirements in the OGTT cluster (Cluster A), and GPT scores perfectly across the parotid cluster on pathogen evidence and care-seeking (Cluster B). \afuagent{} scores perfectly on A2's upward correction and A4's treatment question, but fails the pregnancy-threshold determination required in A1 and A3. In Cluster C, none of the eight systems fully meets C3's requirement to state the lactation contraindication in Chinese labeling for oral single-ingredient dextromethorphan hydrobromide preparations. The case also requires systems to advise the user to check the complete ingredient list and product labeling. Section~\ref{sec:evidence-boundaries} further connects differences to revising earlier unsupported inferences, retaining necessary differential assessment, and checking prerequisites for action. These diagnostic units can guide selection of subsequent training or testing tasks around concrete failures, beyond aggregate rankings.

\paragraph{Evaluation boundaries.}
Simulated dialogues may reflect scenario and expression preferences. Task composition varies across groups, disease and department labels overlap, and some groups are small. A common high effort setting for the seven general-purpose models does not imply equal computation budgets; \afuagent{} scores describe the health assistant service's end-to-end performance on this benchmark. Task scores also depend on medical knowledge, reasoning, and answer completeness, and do not isolate a learning mechanism. Full--Query differences describe score changes between conditions and do not by themselves establish causal effects with all other factors controlled. Cluster-level analyses assess joint quality across related members. Future work can examine stability through comparisons with consistent judgment policies, additional task settings, and repeated runs.

Medical review of cases and rubrics is distinct from human evaluation of model answers. The 90.5\% agreement in Section~\ref{sec:human-judgment-agreement} comes from a sampled check of the original automated decisions, and automated judgments can still err. Broader human review, real interactions, and wider task coverage can further examine the resource's coverage and evaluation reliability.

\Needspace{14\baselineskip}
\section{Conclusion}
\bench{} contributes a medically reviewed resource for evaluating health context learning, with 328 controlled clusters, 1,103 cases, and 4,931 rubric criteria across medical consultation and everyday health management. Shared backgrounds and controlled changes in evidence, patient state, or task goals make related answers comparable; member-specific medical criteria define what each answer must satisfy. Together, these features help evaluate whether systems respond appropriately to those changes and identify where they fall short. Among eight full-context systems, GPT-6 Astra, Claude Opus 5.5, and \afuagent{} rank first through third on both metrics; all eight systems have higher aggregate scores with full context than with query-only inputs in the reported evaluations. Response comparisons still reveal failures in checking historical information, expressing uncertainty when warranted, and ensuring recommendations respect required conditions. The tasks, scoring criteria, and reference results support model comparison and targeted follow-up evaluation.

\Needspace{15\baselineskip}
\begingroup
\raggedright
\setlength{\bibsep}{4pt}
\bibliographystyle{plainnat}
\bibliography{references}
\endgroup
\clearpage
\appendix
\section{Cluster Construction and Complete-Cluster Scores}
\label{sec:case-group-appendix}
This appendix presents three representative clusters and their corresponding results for all eight systems. Appendix~\ref{sec:diffusion-examples} describes each shared context, every member question, and its scoring requirements. Appendix~\ref{sec:paired-responses} reports case scores and geometric cluster scores for the same members in the same order, using matching member identifiers.

\subsection{Construction Examples: Questions, Conditions, and Scoring Requirements}
\label{sec:diffusion-examples}
The following three clusters illustrate diagnostic thresholds, pathogen evidence, and medication ingredients, together with related changes in the requested task. Cluster A contains four members and Clusters B and C each contain three; all ten are shown. Within each cluster, inputs are identical apart from the current question. Member identifiers A1--A4, B1--B3, and C1--C3 match the score tables below and confer no special scoring status. The tables provide faithful English translations of the original Chinese questions, with punctuation and spacing standardized. Scoring requirements summarize each member's independent rubrics.

\paragraph{Cluster A: crossing a diagnostic threshold or preserving the core determination under applicable conditions.}
The shared report gives fasting, 1-hour, and 2-hour glucose values of 4.39, 8.62, and 8.31 mmol/L. The shared context also retains an earlier interpretation using nonpregnant adult criteria and a historical profile marked as the user, male, aged 33. The current questions instead specify a result at 25 weeks of gestation, requiring an updated interpretation and verification of report attribution. A standard 75 g OGTT at \textbf{24--28 weeks of gestation} uses corresponding GDM thresholds of 5.1, 10.0, and 8.5 mmol/L; meeting or exceeding any threshold satisfies the diagnostic criteria~\citep{nhc2023gdm}. The current questions explicitly specify 25 weeks of gestation, within this window, and the rubrics use a standard 75 g OGTT protocol. Responses must still verify the specific testing protocol and report attribution rather than directly applying another person's health profile to this report.

Under these conditions, all three original values fall below the thresholds. The report's elevated flag for the 2-hour value must be interpreted using the appropriate testing protocol rather than directly applying the nonpregnant adult reference limit of 7.8 mmol/L. The response must also verify whose report it is and avoid attributing another person's health information to the pregnant person tested. Table~\ref{tab:case-a} presents the cluster members and their scoring requirements.

\begin{table}[H]
\centering\hcltexttablefont
\caption{Cluster A: corrected OGTT values and a treatment question. The four rows constitute the complete cluster. Fasting and 1-hour values remain 4.39 and 8.62 mmol/L. Clinical interpretation of the threshold comparison is conditional on a verified standard 75 g OGTT at 24--28 weeks of gestation.}
\label{tab:case-a}
\begin{tabularx}{\linewidth}{@{}L{20mm}YY@{}}
\toprule
\rowcolor{hcltexthead}
\textbf{Member} & \textbf{Translated current question} & \textbf{Sample scoring requirements, summarized} \\
\midrule
\textbf{A1}\newline Original values & ``Please analyze this again; these are results from 25 weeks of gestation.'' & Interpret 4.39/8.62/8.31 mmol/L using pregnancy-specific 75 g OGTT criteria and identify that all three are below their thresholds. Distinguish the report's reference range; do not invent a borderline GDM diagnosis solely because a value is close to a threshold. \\
\hcltextrowsep
\textbf{A2}\newline Upward correction & ``The hospital corrected the report. My 2-hour glucose after the glucose load in the OGTT at 25 weeks of gestation was actually 8.81 mmol/L, not the 8.31 reported before. Please analyze it again.'' & Use 8.81 mmol/L, compare it with the 8.5 mmol/L threshold, identify GDM, and recommend obstetric monitoring and management. Do not assume the report belongs to a spouse solely because the profile is male. \\
\hcltextrowsep
\textbf{A3}\newline Downward correction & ``Please analyze this again; these are results from 25 weeks of gestation, and the hospital corrected the report. The glucose value 2 hours after the glucose load was actually 8.21 mmol/L.'' & Use 8.21 rather than 8.31 mmol/L, retaining the determination that the test does not meet GDM criteria. Verify report attribution and continue prenatal follow-up. \\
\hcltextrowsep
\textbf{A4}\newline Treatment question & ``Based on these glucose tolerance results at 25 weeks of gestation, do I need to start insulin immediately?'' & Do not self-initiate or adjust insulin based only on the original values. Compare all three values with their thresholds and have obstetric care determine treatment using subsequent monitoring. \\
\bottomrule
\end{tabularx}
\end{table}

Once the testing protocol and applicable population are verified, correcting 8.31 to 8.81 mmol/L crosses the diagnostic threshold and requires a different core conclusion. Correcting it to 8.21 mmol/L does not cross the threshold but still requires using the updated value. This invariance concerns the core clinical determination, not an unchanged response in every respect. The treatment member retains the original values and changes the response objective. Diagnostic criteria and treatment initiation remain distinct: even when GDM criteria are met, adding insulin depends on whether nutrition, activity, and subsequent glucose management achieve glycemic goals~\citep{ada2026pregnancy}.

\paragraph{Cluster B: pathogen evidence changes management priorities, while task changes alter the response objective.}
The shared background includes swelling and pain of the outer gum below the lower incisors, left-sided palatal pain, and pain on swallowing. These symptoms alone do not establish that parotitis explains all the complaints. The bacterial-evidence member adds aspiration and culture results and asks about management; the viral-evidence member also asks about causes and effects. Their scopes differ, so the evidence comparison focuses on their shared management requirements. Bacterial culture, identification, and antimicrobial susceptibility testing of secretions help inform antibiotic selection. In this example, aspirated pus and a culture positive for Staphylococcus aureus support suppurative bacterial infection requiring physician-directed antibiotics and assessment for accumulated pus~\citep{cheng2026sialadenitis}. A positive mumps-virus IgM result supplies supporting evidence only, requiring clinical assessment and attention to transmission prevention~\citep{cdc2024mumpsserology,cdc2025mumpsclinical}.

\begin{table}[H]
\centering\hcltexttablefont
\caption{Cluster B: bacterial culture, viral serology, and care navigation. The three rows constitute the complete cluster.}
\label{tab:case-b}
\begin{tabularx}{\linewidth}{@{}L{20mm}YY@{}}
\toprule
\rowcolor{hcltexthead}
\textbf{Member} & \textbf{Translated current question} & \textbf{Core response required by the rubrics} \\
\midrule
\textbf{B1}\newline Bacterial evidence & ``I went to the hospital today. They aspirated pus from my parotid gland, and the bacterial culture was positive for Staphylococcus aureus. How should this be managed?'' & Identify suppurative bacterial infection and prioritize antibiotics selected by a clinician using culture or susceptibility results. Reassess recovery and reaccumulation of pus, with further drainage determined by the clinician if needed. \\
\hcltextrowsep
\textbf{B2}\newline Viral evidence & ``I had a blood serology test at the hospital today, and the result was positive for mumps-virus IgM antibodies. What usually causes parotitis? How should it be managed? What effects can it have?'' & Treat positive IgM as supporting evidence and evaluate it alongside symptoms. Explain transmission and complication risks, appropriate isolation, and supportive care; do not recommend antibiotics or drainage routinely. \\
\hcltextrowsep
\textbf{B3}\newline Care navigation & ``Given the swelling and pain in my outer lower gum and the left-sided pain on swallowing, which department should I visit first if they have not improved by Friday?'' & Visit dentistry first if both symptoms persist; consider otolaryngology if throat or palatal pain is the main remaining problem. Seek earlier care for fever, marked facial swelling, or sudden worsening pain. \\
\bottomrule
\end{tabularx}
\end{table}

Pus has already been aspirated in the bacterial-evidence member, making antibiotics and assessment for remaining or recurrent accumulation the priorities rather than unconditional repeat drainage. The viral-evidence member requires infection control and supportive care while preserving diagnostic uncertainty. The care-navigation member introduces neither pathogen test; instead, it turns the shared symptoms into a referral question. Results introduced in the other members must not be transferred to it. The comparison therefore requires identifying both the valid evidence for each question and its current response objective.

\paragraph{Cluster C: breastfeeding medication boundaries under Chinese product labeling.}
The three single-turn questions share the goal of choosing a cough medicine during breastfeeding. They specify an unknown formulation, codeine phosphate, and dextromethorphan as the only cough-suppressant ingredient without alcohol, respectively. This cluster follows Chinese product labeling: the National Medical Products Administration's labeling revision contraindicates oral single-ingredient dextromethorphan hydrobromide preparations during lactation~\citep{nmpa2021dextromethorphan,hainan2022antitussives}. Codeine-containing cough products should also be avoided during breastfeeding because of potential serious adverse effects in infants~\citep{fda2017codeine}.

LactMed's low-milk-exposure information addresses ingredient transfer; it does not override the contraindications of specific Chinese products or justify recommending self-administration~\citep{lactmed2025dextromethorphan}. Since July 1, 2024, China has also classified dextromethorphan, including its salts and single-ingredient preparations, as a Category II psychotropic substance~\citep{nmpa2024dextromethorphan}. This scope must not be generalized to all combination products. Controlled-substance status and lactation contraindications are distinct constraints, both requiring identification of the specific product and applicable rules.

\begin{table}[H]
\centering\hcltexttablefont
\caption{Cluster C: verifying cough preparations during breastfeeding under Chinese labeling. The three rows constitute the complete cluster.}
\label{tab:case-c}
\begin{tabularx}{\linewidth}{@{}L{20mm}YY@{}}
\toprule
\rowcolor{hcltexthead}
\textbf{Member} & \textbf{Translated current question} & \textbf{Core response required by the rubrics} \\
\midrule
\textbf{C1}\newline Unknown formula & ``Can I take cough syrup while breastfeeding?'' & Verify the product name, complete ingredients, and labeling before judging suitability. If it is a Chinese oral single-ingredient dextromethorphan hydrobromide product, identify the lactation contraindication; low milk exposure alone does not justify recommending use. \\
\hcltextrowsep
\textbf{C2}\newline Codeine & ``I'm breastfeeding, and the ingredient list on the bottle includes codeine phosphate. Can I take this cough syrup?'' & Advise against self-administration and recommend consulting a physician or pharmacist about a suitable option. If the product has already been taken and the infant develops unusual drowsiness, feeding difficulty, or abnormal breathing, advise immediate medical care. \\
\hcltextrowsep
\textbf{C3}\newline Dextro\-methorphan & ``I'm breastfeeding and have a cough. I checked the bottle's ingredient list: dextromethorphan is the only cough-suppressant ingredient, and there is no alcohol. Can I take it?'' & State the Chinese lactation contraindication for oral single-ingredient dextromethorphan hydrobromide preparations and advise against self-administration. Verify the complete formula and labeling to determine whether this restriction applies to the user's product. Alcohol absence, infant characteristics, or overseas milk-transfer data do not remove the contraindication; do not directly provide dosing or duration, or advise self-directed breastfeeding cessation. \\
\bottomrule
\end{tabularx}
\end{table}

This cluster tests whether new ingredient information is used to identify applicable medication boundaries. Dextromethorphan and the absence of alcohol do not provide sufficient grounds for approving use; an ``only cough-suppressant ingredient'' also does not establish that the full product has only one active ingredient. Responses must use the available facts, identify unverified product conditions, and state the applicable labeling requirements, without treating evidence from different jurisdictions as interchangeable medication advice.

\FloatBarrier
\subsection{Eight-System Scores for the Same Clusters}
\label{sec:paired-responses}
The A--C groups in Table~\ref{tab:case-cluster-scores} correspond to the same clusters in Tables~\ref{tab:case-a}--\ref{tab:case-c} and report 80 member scores and 24 geometric cluster scores for all eight systems under full context. A1--A4, B1--B3, and C1--C3 map directly to the question tables, and each cluster includes every member retained in the final benchmark. All scores use a 100-point scale. Following Section~\ref{sec:scoring}, the geometric cluster score $G$ is calculated from the final case scores rounded to four decimal places, then converted to the 100-point scale; displayed values use two decimal places.

% Generated by analysis/revision_20261110/render_assets.py.
\begin{table}[H]
\centering\hcldensetablefont
\caption{All member and geometric cluster scores for case clusters A--C (full context; 0--100 scale).}
\label{tab:case-cluster-scores}
\label{tab:case-a-scores}
\label{tab:case-b-scores}
\label{tab:case-c-scores}
\begin{tabularx}{\linewidth}{@{}L{24mm}|*{8}{>{\centering\arraybackslash}X}@{}}
\toprule
\rowcolor{hcltablehead}
Member / cluster & \shortstack{GPT-6\\Astra} & \shortstack{Claude\\Opus 5.5} & \afuagent{} & \shortstack{Qwen3.8-\\27B} & GLM-5.3 & Kimi K3 & \shortstack{DeepSeek-\\V4.1-Flash} & \shortstack{Gemini\\3.8 Flash} \\
\midrule
\rowcolor{hcltableband}
\multicolumn{9}{@{}l@{}}{\textbf{Case cluster A}} \\
\rowcolor{hcltablerow}
A1 & \textbf{100.00} & \textbf{100.00} & 0.00 & 66.67 & \textbf{100.00} & 66.67 & \textbf{100.00} & \textbf{100.00} \\ % key=A1
A2 & \textbf{100.00} & \textbf{100.00} & \textbf{100.00} & \textbf{100.00} & \textbf{100.00} & \textbf{100.00} & \textbf{100.00} & \textbf{100.00} \\ % key=A2
\rowcolor{hcltablerow}
A3 & 55.56 & \textbf{100.00} & 0.00 & 55.56 & \textbf{100.00} & 55.56 & 55.56 & 55.56 \\ % key=A3
A4 & \textbf{100.00} & \textbf{100.00} & \textbf{100.00} & \textbf{100.00} & \textbf{100.00} & \textbf{100.00} & 55.56 & \textbf{100.00} \\ % key=A4
\rowcolor{hcltablerow}
$G_{\mathrm{A}}$ & 86.34 & \textbf{100.00} & 0.00 & 78.01 & \textbf{100.00} & 78.01 & 74.54 & 86.34 \\ % key=A.G
\midrule
\rowcolor{hcltableband}
\multicolumn{9}{@{}l@{}}{\textbf{Case cluster B}} \\
\rowcolor{hcltablerow}
B1 & \textbf{100.00} & 71.43 & 0.00 & 71.43 & 71.43 & \textbf{100.00} & \textbf{100.00} & 0.00 \\ % key=B1
B2 & \textbf{100.00} & 66.67 & 33.33 & \textbf{100.00} & 33.33 & 33.33 & 66.67 & 33.33 \\ % key=B2
\rowcolor{hcltablerow}
B3 & \textbf{100.00} & \textbf{100.00} & \textbf{100.00} & 42.86 & 0.00 & \textbf{100.00} & 42.86 & 42.86 \\ % key=B3
$G_{\mathrm{B}}$ & \textbf{100.00} & 78.09 & 0.00 & 67.40 & 0.00 & 69.33 & 65.87 & 0.00 \\ % key=B.G
\midrule
\rowcolor{hcltableband}
\multicolumn{9}{@{}l@{}}{\textbf{Case cluster C}} \\
\rowcolor{hcltablerow}
C1 & 47.83 & \textbf{82.61} & 0.00 & 65.22 & 39.13 & 0.00 & 39.13 & 47.83 \\ % key=C1
C2 & \textbf{100.00} & \textbf{100.00} & \textbf{100.00} & \textbf{100.00} & \textbf{100.00} & \textbf{100.00} & 0.00 & \textbf{100.00} \\ % key=C2
\rowcolor{hcltablerow}
C3 & 0.00 & 0.00 & 0.00 & 0.00 & 0.00 & 0.00 & 0.00 & 0.00 \\ % key=C3
$G_{\mathrm{C}}$ & 0.00 & 0.00 & 0.00 & 0.00 & 0.00 & 0.00 & 0.00 & 0.00 \\ % key=C.G
\bottomrule
\end{tabularx}
\par\vspace{3pt}
\begin{minipage}{\linewidth}\fontsize{8.5}{9.6}\selectfont\raggedright Member order follows Tables~\ref{tab:case-a}--\ref{tab:case-c}; $G$ uses complete clusters. The seven general-purpose models use thinking/high; \afuagent{} is evaluated using Ant Group's AI health assistant service. Bold marks row maxima among positive scores, including ties.
\end{minipage}
\end{table}

\paragraph{Cluster A: updating values and verifying report attribution.}
Claude and GLM score 100.00 on all four members and therefore have geometric cluster scores of 100.00. GPT scores 100.00 on A1, A2, and A4, but 55.56 on A3, reducing its cluster score to 86.34. In A3, its P1 requirement to verify attribution between the male profile and the pregnancy report is judged unmet; this necessary contextual check accounts for the deduction. \afuagent{} scores perfectly on A2 and A4, but fails the P0 requirement on pregnancy-specific thresholds and the determination that GDM criteria are not met on both A1 and A3, yielding a zero cluster score. The cluster jointly tests corrected values, threshold interpretation, and attribution within a shared setting, showing that multiple contextual requirements must be satisfied together.

\paragraph{Cluster B: pathogen evidence and care-seeking goals impose different requirements.}
GPT scores 100.00 on B1--B3. Claude has a geometric cluster score of 78.09, followed by Kimi, Qwen, and DeepSeek at 69.33, 67.40, and 65.87. Claude scores 71.43 on B1: the final judgment passes bacterial-infection identification but retains the P1 failure on follow-up after aspiration and assessment for recurrent pus accumulation. \afuagent{}, GLM, and Gemini each have a zero-scoring member and hence a zero cluster score. GLM's zero on B3, for example, results from P1 failures concerning department-selection branches and prompt care if symptoms worsen, which reduce the net score to zero. Against related symptom backgrounds, pathogen interpretation, management, and care-seeking goals must each be addressed using the evidence available to the current member.

\paragraph{Cluster C: interpreting ingredients under applicable product labeling.}
All seven systems other than DeepSeek score 100.00 on C2. On C3, all eight fail the P0 requirement to state the lactation contraindication in Chinese labeling for single-ingredient oral dextromethorphan hydrobromide products, yielding zero C3 scores and zero three-member geometric scores. General caution in a response differs from this specific requirement: DeepSeek, for example, already advises against use, but describes caution or avoidance rather than explicitly stating the required contraindication. This cluster exposes a shared weakness in combining new ingredient information with the applicable product-specific requirements across closely related medication questions.
\FloatBarrier

\section{Label, Coverage, and Subgroup Result Details}
\label{sec:benchmark-details}
This appendix supplements the label boundaries in Section~\ref{sec:taxonomy} and the distributions in Section~\ref{sec:distribution}. The scenario and difficulty groups in Table~\ref{tab:scenarios} connect task categories and context requirements to practical evaluation uses; the details below preserve the finer definitions, coverage denominators, and detailed subgroup results.

\subsection{Difficulty Subcategories and Label Distributions}
\label{sec:difficulty-details}
Table~\ref{tab:difficulty-definitions} defines the 14 subcategories within the four context difficulty families. A case may have several difficulties; its primary difficulty denotes the most decisive requirement for the current task. For example, D1 in the main text has both Complex and Fragmented Context labels, with Complex Context as primary; its illustration is not an exclusive classification rule. These labels describe input structure, not errors already made by a model.
\begingroup
\hcltexttablefont
\setlength{\LTleft}{\fill}
\setlength{\LTright}{\fill}
\begin{longtable}{@{}L{0.30\linewidth}L{\dimexpr0.70\linewidth-2\tabcolsep\relax}@{}}
\caption{Definitions of the fourteen context-difficulty subcategories, grouped into four categories. A case may exhibit multiple difficulties.}\label{tab:difficulty-definitions}\\
\toprule
\rowcolor{hcltexthead}
\textbf{Difficulty subcategory} & \textbf{Definition} \\
\midrule
\endfirsthead
\multicolumn{2}{@{}l@{}}{Context-difficulty subcategory definitions (continued)}\\
\toprule
\rowcolor{hcltexthead}
\textbf{Difficulty subcategory} & \textbf{Definition} \\
\midrule
\endhead
\midrule
\multicolumn{2}{r}{Continued on next page}\\
\endfoot
\bottomrule
\endlastfoot
\rowcolor{hcltextband}
\multicolumn{2}{@{}l@{}}{\textbf{Complex Context}}\\*
Multiple health subjects & The context concerns multiple individuals with distinct health information, requiring symptoms, histories, tests, or medications to be assigned to the correct person. \\
\hcltextrowsep
Interleaved topics & Multiple health topics run in parallel, alternate, pause, or recur, requiring the relevant topic thread to be tracked across the conversation. \\
\hcltextrowsep
Reference and ellipsis dependence & Pronouns, demonstratives, ellipsis, or continuation expressions depend on earlier context to recover the intended person, measurement, medication, test, event, or action. \\
\hcltextrowsep
Multiple questions in one query & A single query contains at least two questions, judgment goals, or task requirements that must be addressed separately. \\
\hcltextrowsep
Dynamic context updates & Later information explicitly corrects, changes, or supersedes an earlier fact, health state, plan, or conclusion, requiring identification of the currently valid information. \\
\rowcolor{hcltextband}
\multicolumn{2}{@{}l@{}}{\textbf{Noisy Context}}\\*
Colloquial expressions and terminology variants & Key health information is expressed through colloquial or dialectal language, informal names, abbreviations, mixed languages, or nonstandard terminology, requiring interpretation of its intended meaning. \\
\hcltextrowsep
Input noise & The input contains typographical errors, homophone substitutions, omissions, segmentation errors, or formatting irregularities, while the intended meaning remains uniquely recoverable from the available context. \\
\hcltextrowsep
Irrelevant content & The facts needed for the answer are embedded in small talk, digressions, or other irrelevant material whose removal leaves the relevant facts and answer unchanged. \\
\rowcolor{hcltextband}
\multicolumn{2}{@{}l@{}}{\textbf{Fragmented Context}}\\*
Longitudinal records & Repeated records of the same type of health measurement, symptom, diet, exercise, sleep, or medication span multiple time points and must be integrated to assess their trajectory. \\
\hcltextrowsep
Event timeline & Events such as tests, diagnoses, medication use, or symptom changes occur at different times, and their order, stage, duration, or intervals affect the answer. \\
\hcltextrowsep
Evidence fragmented across turns & A complete answer requires complementary facts distributed across multiple dialogue turns, with no single turn providing sufficient evidence on its own. \\
\rowcolor{hcltextband}
\multicolumn{2}{@{}l@{}}{\textbf{Conflicting Context}}\\*
Conflicting facts and interpretations & Factual statements or interpretations about the same person, time, conditions, and health proposition cannot simultaneously hold, with no explicit update resolving the conflict. \\
\hcltextrowsep
Conflicting applicability conditions & The same health information receives inconsistent interpretations under different populations, sampling conditions, units, measurement methods, disease stages, or reference criteria. \\
\hcltextrowsep
Conflicting standards and recommendations & The same health problem at the same time is associated with reference standards, management targets, or recommended actions that cannot be directly adopted together, without an explicit indication that one supersedes another. \\
\end{longtable}
\endgroup

Main-text Table~\ref{tab:label-distribution}(b) provides exact counts corresponding to the main-text distribution figure. Medical scenarios and primary difficulties are each mutually exclusive, with all 1,103 cases as the denominator. Multilabel difficulty coverage instead counts distinct cases within each category: a case may enter several categories, so these counts do not sum to the dataset size.

\subsection{Disease and Department Coverage}
Disease coverage statistics include 924 cases involving disease-related content, spanning 42 disease categories and Other Diseases. Department coverage statistics include 785 cases involving department-related information, spanning 25 department categories and Other Departments. The left and right panels of Table~\ref{tab:clinical-distribution} report the eight most frequent categories and the corresponding Other category. Percentages use the department-related and disease-related case sets, respectively, as their denominators.

Among disease-related cases, diabetes covers 184 cases (19.91\%), hypertension 155 (16.77\%), and bacterial infection 128 (13.85\%). Among department-related cases, endocrinology covers 229 cases (29.17\%), dermatology and venereology 209 (26.62\%), and gastroenterology 205 (26.11\%). A case can involve several diseases or departments.
% Structurally consolidated from the 2026-10-28 source snapshot; no scores recalculated.
\begin{table}[H]
\centering\hcldensetablefont
\caption{Department and disease coverage: the eight most frequent categories and Other in each dimension.}
\label{tab:clinical-distribution}
\label{tab:department-distribution}
\label{tab:disease-distribution}
\begin{tabularx}{\linewidth}{@{}Yrr@{\hspace{6mm}}Yrr@{}}
\toprule
\rowcolor{hcltablehead}
\multicolumn{3}{c}{\textbf{Departments (785 related cases)}} & \multicolumn{3}{c}{\textbf{Diseases (924 related cases)}} \\
\cmidrule(lr){1-3}\cmidrule(lr){4-6}
\rowcolor{hcltableband}
Category & Cases & Share (\%) & Category & Cases & Share (\%) \\
\midrule
\rowcolor{hcltablerow}
Endocrinology & 229 & 29.17 & Diabetes & 184 & 19.91 \\
Dermatology and Venereology & 209 & 26.62 & Hypertension & 155 & 16.77 \\
\rowcolor{hcltablerow}
Gastroenterology & 205 & 26.11 & Bacterial Infection & 128 & 13.85 \\
Cardiovascular Medicine & 181 & 23.06 & Common Cold & 116 & 12.55 \\
\rowcolor{hcltablerow}
Infectious Diseases & 168 & 21.40 & Anemia & 98 & 10.61 \\
Respiratory Medicine & 141 & 17.96 & Eczema & 94 & 10.17 \\
\rowcolor{hcltablerow}
Hematology & 108 & 13.76 & Fungal Infection & 76 & 8.23 \\
Otolaryngology & 80 & 10.19 & Dyspepsia & 70 & 7.58 \\
\rowcolor{hcltablerow}
Other Departments & 339 & 43.18 & Other Diseases & 912 & 98.70 \\
\bottomrule
\end{tabularx}
\par\vspace{3pt}
\begin{minipage}{\linewidth}\fontsize{8.5}{9.6}\selectfont\raggedright Percentages use 785 department-related cases and 924 disease-related cases as their respective denominators; a case may enter multiple categories.
\end{minipage}
\end{table}

Other Diseases and Other Departments capture related content outside the listed disease and department categories and can co-occur with specific categories. Figure~\ref{fig:clinical-coverage-pies} presents the composition of category occurrences: disease-related cases contribute 2,569 occurrences, and department-related cases contribute 2,288. The remaining 34 disease categories and 17 department categories are grouped together in the respective panels.
\input{figures/clinical_coverage_pies}

Disease and department labels support content-based retrieval and subgroup comparisons; their frequencies do not estimate population prevalence, diagnosis rates, or actual clinical visits. They describe dimensions separate from scenarios and context difficulties. For example, a medication-choice task can be further filtered by disease content, while the same disease can occur in explanation, assessment, or care-triage tasks.

\subsection{Detailed Subgroup Results}
\subsubsection{Subgroup Results with Full Context}
\label{sec:full-category-results}
Tables~\ref{tab:scenario-results} and~\ref{tab:clinical-results} supplement the main-text analysis with eight-system case means by medical scenario, primary context difficulty, department, and disease. Scenarios and primary difficulties are each mutually exclusive; department and disease groups can overlap. Cases are equally weighted within a group, using the scoring function and final judgments described in Section~\ref{sec:scoring}. The $N$ column reports group sizes.
% Generated by analysis/revision_20261110/render_assets.py.
\begin{table}[H]
\centering\hcldensetablefont
\caption{Case means by medical scenario and primary context difficulty (full context; 0--100 scale).}
\label{tab:scenario-results}
\label{tab:difficulty-results}
\begin{tabularx}{\linewidth}{@{}L{27mm}>{\centering\arraybackslash}p{6mm}|*{8}{>{\centering\arraybackslash}X}@{}}
\toprule
\rowcolor{hcltablehead}
Category & $N$ & \shortstack{GPT-6\\Astra} & \shortstack{Claude\\Opus 5.5} & \afuagent{} & \shortstack{Qwen3.8-\\27B} & GLM-5.3 & Kimi K3 & \shortstack{DeepSeek-\\V4.1-Flash} & \shortstack{Gemini\\3.8 Flash} \\
\midrule
\rowcolor{hcltableband}
\multicolumn{10}{@{}l@{}}{\textbf{Medical scenario}} \\
\rowcolor{hcltablerow}
Medical Explanation & 41 & \textbf{73.22} & 58.84 & 59.84 & 51.41 & 55.48 & 43.40 & 44.44 & 42.09 \\ % key=scenario.0
Health Assessment & 498 & \textbf{73.52} & 68.83 & 61.81 & 57.45 & 50.91 & 45.94 & 44.25 & 40.47 \\ % key=scenario.1
\rowcolor{hcltablerow}
Clinical Decision-Making & 273 & \textbf{76.42} & 63.88 & 60.62 & 56.06 & 51.01 & 45.16 & 46.74 & 45.29 \\ % key=scenario.2
Care Triage & 127 & \textbf{81.84} & 80.38 & 70.88 & 66.96 & 63.56 & 64.16 & 55.46 & 52.12 \\ % key=scenario.3
\rowcolor{hcltablerow}
Health Management & 164 & \textbf{72.50} & 66.48 & 58.49 & 52.93 & 49.08 & 50.67 & 41.22 & 38.53 \\ % key=scenario.4
\midrule
\rowcolor{hcltableband}
\multicolumn{10}{@{}l@{}}{\textbf{Primary context difficulty}} \\
\rowcolor{hcltablerow}
Complex Context & 545 & \textbf{74.96} & 66.85 & 63.68 & 55.86 & 50.06 & 46.25 & 47.06 & 43.78 \\ % key=primary_difficulty.0
Fragmented Context & 305 & \textbf{72.19} & 66.38 & 54.86 & 54.40 & 50.80 & 45.72 & 38.24 & 36.35 \\ % key=primary_difficulty.1
\rowcolor{hcltablerow}
Conflicting Context & 81 & \textbf{75.82} & 66.24 & 55.78 & 53.06 & 50.83 & 45.84 & 41.54 & 37.16 \\ % key=primary_difficulty.2
Noisy Context & 65 & \textbf{79.21} & 77.60 & 72.83 & 67.66 & 64.15 & 61.88 & 52.28 & 54.05 \\ % key=primary_difficulty.3
\rowcolor{hcltablerow}
Other Difficulties & 107 & \textbf{80.38} & 76.17 & 71.83 & 69.86 & 61.77 & 61.31 & 59.31 & 53.33 \\ % key=primary_difficulty.4
\bottomrule
\end{tabularx}
\par\vspace{3pt}
\begin{minipage}{\linewidth}\fontsize{8.5}{9.6}\selectfont\raggedright $N$ is the case count; each dimension partitions all 1,103 cases. Other Difficulties covers cases outside the four named categories. The seven general-purpose models use thinking/high; \afuagent{} is evaluated using Ant Group's AI health assistant service. Bold marks row maxima among positive scores, including ties.
\end{minipage}
\end{table}

% Generated by analysis/revision_20261110/render_assets.py.
\begin{table}[H]
\centering\hcldensetablefont
\caption{Case means by department and disease (full context; 0--100 scale).}
\label{tab:clinical-results}
\label{tab:department-results}
\label{tab:disease-results}
\begin{tabularx}{\linewidth}{@{}L{27mm}>{\centering\arraybackslash}p{6mm}|*{8}{>{\centering\arraybackslash}X}@{}}
\toprule
\rowcolor{hcltablehead}
Category & $N$ & \shortstack{GPT-6\\Astra} & \shortstack{Claude\\Opus 5.5} & \afuagent{} & \shortstack{Qwen3.8-\\27B} & GLM-5.3 & Kimi K3 & \shortstack{DeepSeek-\\V4.1-Flash} & \shortstack{Gemini\\3.8 Flash} \\
\midrule
\rowcolor{hcltableband}
\multicolumn{10}{@{}l@{}}{\textbf{Department}} \\
\rowcolor{hcltablerow}
Endocrinology & 229 & 73.61 & \textbf{74.54} & 67.24 & 62.32 & 58.48 & 52.63 & 47.27 & 44.07 \\ % key=department_labels.0
Dermatology and Venereology & 209 & \textbf{77.56} & 70.09 & 66.40 & 57.22 & 53.43 & 50.86 & 45.57 & 45.86 \\ % key=department_labels.1
\rowcolor{hcltablerow}
Gastroenterology & 205 & \textbf{71.82} & 68.52 & 61.42 & 56.40 & 52.83 & 49.20 & 44.69 & 42.56 \\ % key=department_labels.2
Cardiovascular Medicine & 181 & \textbf{71.24} & 66.84 & 61.87 & 57.92 & 53.18 & 48.63 & 45.51 & 41.96 \\ % key=department_labels.3
\rowcolor{hcltablerow}
Infectious Diseases & 168 & \textbf{74.38} & 70.64 & 66.20 & 60.03 & 55.01 & 51.84 & 45.51 & 47.26 \\ % key=department_labels.4
Respiratory Medicine & 141 & \textbf{74.55} & 62.11 & 61.06 & 47.87 & 48.70 & 43.89 & 43.98 & 41.27 \\ % key=department_labels.5
\rowcolor{hcltablerow}
Hematology & 108 & \textbf{70.84} & 62.93 & 58.79 & 53.75 & 49.25 & 44.05 & 42.02 & 34.92 \\ % key=department_labels.6
Otolaryngology & 80 & \textbf{75.32} & 66.15 & 59.91 & 47.97 & 47.81 & 44.12 & 38.55 & 37.21 \\ % key=department_labels.7
\rowcolor{hcltablerow}
Other Departments & 339 & \textbf{71.93} & 63.57 & 59.91 & 54.11 & 48.93 & 44.41 & 42.75 & 41.10 \\ % key=department_labels.8
\midrule
\rowcolor{hcltableband}
\multicolumn{10}{@{}l@{}}{\textbf{Disease}} \\
\rowcolor{hcltablerow}
Diabetes & 184 & 71.81 & \textbf{73.17} & 67.50 & 60.57 & 56.83 & 51.38 & 44.92 & 42.84 \\ % key=disease_labels.0
Hypertension & 155 & \textbf{70.64} & 64.85 & 60.56 & 59.21 & 53.27 & 48.99 & 45.17 & 40.99 \\ % key=disease_labels.1
\rowcolor{hcltablerow}
Bacterial Infection & 128 & \textbf{72.97} & 67.67 & 62.22 & 55.82 & 54.53 & 51.47 & 44.43 & 45.36 \\ % key=disease_labels.2
Common Cold & 116 & \textbf{75.60} & 59.61 & 59.50 & 47.05 & 47.44 & 40.89 & 41.62 & 39.92 \\ % key=disease_labels.3
\rowcolor{hcltablerow}
Anemia & 98 & \textbf{71.05} & 62.15 & 58.47 & 52.86 & 49.06 & 42.72 & 42.88 & 34.32 \\ % key=disease_labels.4
Eczema & 94 & \textbf{78.42} & 71.42 & 67.47 & 56.08 & 54.52 & 48.33 & 43.41 & 48.27 \\ % key=disease_labels.5
\rowcolor{hcltablerow}
Fungal Infection & 76 & \textbf{77.46} & 74.98 & 72.66 & 63.74 & 55.85 & 52.75 & 47.85 & 51.22 \\ % key=disease_labels.6
Dyspepsia & 70 & \textbf{68.61} & 64.43 & 55.82 & 54.85 & 56.31 & 46.18 & 45.44 & 40.30 \\ % key=disease_labels.7
\rowcolor{hcltablerow}
Other Diseases & 912 & \textbf{74.89} & 67.82 & 62.20 & 56.23 & 50.93 & 46.88 & 44.78 & 42.20 \\ % key=disease_labels.8
\bottomrule
\end{tabularx}
\par\vspace{3pt}
\begin{minipage}{\linewidth}\fontsize{8.5}{9.6}\selectfont\raggedright $N$ is the case count. Each dimension shows its eight most frequent categories and Other; a case may have multiple categories. The seven general-purpose models use thinking/high; \afuagent{} is evaluated using Ant Group's AI health assistant service. Bold marks row maxima among positive scores, including ties.
\end{minipage}
\end{table}

\subsubsection{Full-Context versus Query-Only Subgroup Comparisons}
\label{sec:ablation-detail-results}
Table~\ref{tab:context-ablation-results} reports paired Full$-$Query differences for all eight systems by medical scenario, primary difficulty, and user-turn count. Each difference compares the same cases: positive values indicate a higher group case mean with full context, and negative values a higher mean with the query only. The interpretation follows Section~\ref{sec:context-ablation}.
% Generated by analysis/revision_20261110_afu_query/render_assets.py.
\begin{table}[H]
\centering\hcldensetablefont
\caption{Full$-$Query differences for all eight systems by scenario, primary difficulty, and user turns (points).}
\label{tab:context-ablation-results}
\label{tab:context-ablation-difficulty}
\label{tab:context-ablation-turns}
\begin{tabularx}{\linewidth}{@{}L{27mm}>{\centering\arraybackslash}p{6mm}|*{8}{>{\centering\arraybackslash}X}@{}}
\toprule
\rowcolor{hcltablehead}
Category & $N$ & \shortstack{GPT-6\\Astra} & \shortstack{Claude\\Opus 5.5} & \afuagent{} & \shortstack{Qwen3.8-\\27B} & GLM-5.3 & Kimi K3 & \shortstack{DeepSeek-\\V4.1-Flash} & \shortstack{Gemini\\3.8 Flash} \\
\midrule
\rowcolor{hcltableband}
\multicolumn{10}{@{}l@{}}{\textbf{Medical scenario}} \\
\rowcolor{hcltablerow}
Medical Explanation & 41 & +28.43 & +24.48 & +17.77 & +17.11 & +24.44 & +14.63 & +9.96 & +18.31 \\ % key=scenario.0
Health Assessment & 498 & +30.29 & +33.95 & +23.10 & +17.69 & +18.72 & +16.48 & +12.08 & +11.45 \\ % key=scenario.1
\rowcolor{hcltablerow}
Clinical Decision-Making & 273 & +32.68 & +25.17 & +20.94 & +14.30 & +16.02 & +13.31 & +15.50 & +11.50 \\ % key=scenario.2
Care Triage & 127 & +27.05 & +35.63 & +23.98 & +17.10 & +19.81 & +22.01 & +12.09 & +12.55 \\ % key=scenario.3
\rowcolor{hcltablerow}
Health Management & 164 & +33.17 & +34.86 & +22.02 & +17.87 & +20.22 & +20.09 & +11.10 & +11.34 \\ % key=scenario.4
\midrule
\rowcolor{hcltableband}
\multicolumn{10}{@{}l@{}}{\textbf{Primary context difficulty}} \\
\rowcolor{hcltablerow}
Complex Context & 545 & +33.97 & +33.56 & +25.67 & +19.09 & +19.36 & +16.30 & +16.36 & +16.03 \\ % key=primary_difficulty.0
Fragmented Context & 305 & +33.65 & +34.45 & +20.42 & +17.80 & +21.39 & +18.78 & +10.52 & +9.62 \\ % key=primary_difficulty.1
\rowcolor{hcltablerow}
Conflicting Context & 81 & +33.88 & +34.19 & +18.56 & +18.43 & +21.75 & +23.57 & +11.76 & +10.90 \\ % key=primary_difficulty.2
Noisy Context & 65 & +20.30 & +22.91 & +23.04 & +9.90 & +14.09 & +16.23 & +4.27 & +8.45 \\ % key=primary_difficulty.3
\rowcolor{hcltablerow}
Other Difficulties & 107 & +11.28 & +18.42 & +12.94 & +5.12 & +7.22 & +8.96 & +6.19 & -0.54 \\ % key=primary_difficulty.4
\midrule
\rowcolor{hcltableband}
\multicolumn{10}{@{}l@{}}{\textbf{User turns (including current query)}} \\
\rowcolor{hcltablerow}
1 & 157 & +21.11 & +29.02 & +21.53 & +19.45 & +26.42 & +24.07 & +16.76 & +16.74 \\ % key=turn_bin.0
2--5 & 499 & +33.22 & +33.66 & +24.12 & +21.49 & +22.16 & +20.16 & +16.90 & +11.18 \\ % key=turn_bin.1
\rowcolor{hcltablerow}
6--10 & 234 & +33.66 & +31.89 & +25.47 & +13.91 & +12.96 & +13.24 & +11.19 & +13.00 \\ % key=turn_bin.2
11--20 & 139 & +33.31 & +26.53 & +17.75 & +12.48 & +11.91 & +6.93 & +7.17 & +11.54 \\ % key=turn_bin.3
\rowcolor{hcltablerow}
21--40 & 56 & +14.56 & +23.95 & -1.16 & -11.97 & +4.01 & +0.17 & -14.97 & -2.13 \\ % key=turn_bin.4
$>40$ & 18 & +46.37 & +65.74 & +45.76 & +23.50 & +22.60 & +34.61 & +9.77 & +17.33 \\ % key=turn_bin.5
\bottomrule
\end{tabularx}
\par\vspace{3pt}
\begin{minipage}{\linewidth}\fontsize{8.5}{9.6}\selectfont\raggedright $N$ is the paired-case count; each dimension covers 1,103 cases. Differences use unrounded within-group means; positive values favor Full, negative values Query. The seven general-purpose models use thinking/high; \afuagent{} is evaluated using Ant Group's AI health assistant service. Differences are descriptive comparisons. 
\end{minipage}
\end{table}

\FloatBarrier

\end{document}